\documentclass[10pt,twoside]{article}

\usepackage{adjustbox}
\usepackage{amsmath}
\usepackage{amssymb}
\usepackage{caption}
\usepackage{fancyhdr}
\usepackage{float}
\usepackage[T1]{fontenc}
\usepackage{graphicx}
\usepackage{hhline}
\usepackage[utf8]{inputenc}
\usepackage{mathtools}
\usepackage{multicol}
\usepackage{multirow}
\usepackage{subcaption}
\usepackage{wasysym}
\usepackage[svgnames]{xcolor}
\usepackage[paperheight=27.94cm,paperwidth=21.59cm,left=3.17cm,right=3.17cm,top=2.54cm,bottom=2.54cm]{geometry}
\usepackage[hidelinks]{hyperref}

\begin{document}

\vspace{1\baselineskip}
\subsection*{Original Paper}

Chen Lin\textsuperscript{1}; Safoora Yousefi\textsuperscript{1}, PhD; Elvis Kahoro\textsuperscript{2}; Donghai Liang\textsuperscript{3}, PhD; Jeremy A. Sarnat\textsuperscript{3}, ScD; Eugene Agichtein\textsuperscript{1}, PhD  

\textsuperscript{1}Department of Computer Science, Emory University, Atlanta, GA, United States

\textsuperscript{2}Department of Computer Science, Pomona College, Claremont, CA, United States

\textsuperscript{3}Gangarosa Department of Environmental Health, Rollins School of Public Health, Emory University, Atlanta, GA, United States

\vspace{1\baselineskip}
\textbf{Corresponding Author:}

Chen Lin, 

Department of Computer Science

Emory University 

MSC Building, W302 \\ 400 Dowman Drive \\ Atlanta, GA 30322

\vspace{1\baselineskip}
\section*{Detecting Elevated Air Pollution Levels by Monitoring Web Search Queries: Deep Learning-Based Time Series Forecasting}

\subsection*{Abstract}

\textbf{Background:}

Real-time air pollution monitoring is a valuable tool for public health and environmental surveillance. In recent years, there has been a dramatic increase in air pollution forecasting and monitoring research using artificial neural networks (ANNs). Most of the prior work relied on modeling pollutant concentrations collected from ground-based monitors and meteorological data for long-term forecasting of outdoor ozone, oxides of nitrogen, and PM\textsubscript{2.5}. Given that traditional, highly sophisticated air quality monitors are expensive and are not universally available, these models cannot adequately serve those not living near pollutant monitoring sites. Furthermore, because prior models were built on physical measurement data collected from sensors, they may not be suitable for predicting public health effects experienced from pollution exposure.

\textbf{Objective:}

This study aims to develop and validate models to ``nowcast" the observed pollution levels using Web search data, which is publicly available in near real-time from major search engines.

\textbf{Methods:}

We developed novel machine learning-based models using both traditional supervised classification methods and state-of-the-art deep learning methods to detect elevated air pollution levels at the US city level, by using generally available meteorological data and aggregate Web-based search volume data derived from Google Trends. We validated the performance of these methods by predicting three critical air pollutants (ozone (O\textsubscript{3}), nitrogen dioxide (NO\textsubscript{2}), and fine particulate matter (PM\textsubscript{2.5})), across ten major U.S. metropolitan statistical areas (MSAs) in 2017 and 2018. We also explore different variations of the long-short term memory (LSTM) model and propose a novel search term Dictionary Learner-Long-Short Term Memory (DL-LSTM) model to learn sequential patterns across multiple search terms for prediction.

\textbf{Results:} 

The top-performing model was a deep neural sequence model LSTM, using meteorological and Web search data, and reached an accuracy of 0.82 (F1 score: 0.51) for O\textsubscript{3,} 0.74 (F1 score:0.41) for NO\textsubscript{2}, and 0.85 (F1 score: 0.27) for PM\textsubscript{2.5}, when used for detecting elevated pollution levels. Compared with only using meteorological data, the proposed method achieved superior accuracy by incorporating Web search data. 

\textbf{Conclusions:}

The results show that incorporating Web search data with meteorological data improves nowcasting performance for all three pollutants and suggest promising novel applications for tracking global physical phenomena using Web search data.

\vspace{1\baselineskip}
\textbf{Keywords: }

Nowcasting of air pollution; online public health surveillance; neural network sequence modeling; search engine log analysis; Air pollution exposure assessment

\vspace{1\baselineskip}


\subsection*{Introduction}

Online crowd surveillance has been used as a means of tracking emergent risk to public health [1-3]. Most commonly, these efforts involve the collection of online search queries to document acute changes in incidence or symptoms occurrence to primary infectious disease agents, such as influenza [4-7], Ebola [8], dengue fever [9], and COVID-19 [10]. These methods have the potential to provide public health and medical professionals with benefits over traditional health surveillance and environmental epidemiology in their ability to capture both personal exposures and response dynamics at more sensitive spatial and temporal scales [11].

\vspace{1\baselineskip}
Despite the promise of these approaches on infectious diseases, only a limited number of studies have examined how crowd-surveillance approaches can be used to track environmental exposures and, less frequently, responses to non-infectious environmental-mediated disease processes [12-14]. The global burden of disease attributable to outdoor and indoor air pollution has been quantified by recent efforts and has increased public awareness on the severity of this public health crisis worldwide [15]. Urban air pollution, therefore, provides a key test case for the evaluation of online surveillance approaches for non-infectious environmental risks. The online surveillance approach is distinct from traditional approaches for measuring urban air pollution exposures. Therefore, it could possibly serve as a substitute or complement to existing approaches. Traditional indicators of air pollution exposures, namely concentrations measured at ambient monitoring sites, are widely used to assess health effects associated with air pollution in epidemiological studies. However, the use of ambient monitoring measurements as surrogates of exposure may result in misclassification of health response and potential risk, especially for those not living near pollutant monitoring sites [16-18]. Moreover, ambient monitoring, by design, provides information on measured outdoor pollutant concentrations, and may not necessarily reflect accurate personal exposures for individuals spending the majority of their time indoors, or for those with preexisting biological susceptibility to air pollution. Several recent studies have focused on using smartphones within distributed air pollution sensing networks, where users record and upload local air pollution conditions to crowd-generated, geospatially-refined pollution maps [12-14]. These studies demonstrate the feasibility of online crowd-generated participation in projects predicated on urban air pollution awareness.

\vspace{1\baselineskip}
To our knowledge, few studies have investigated the feasibility of using Web search data to produce accurate ``nowcasts" of urban air pollution levels in real-time. Conducting accurate predictions using Web search data is a challenging task with two major challenges. The first is the selection of search terms to capture people’s responses comprehensively. Several approaches have been proposed to select the search terms. For example, some studies preliminarily prepare keywords related to the target disease and then use these keywords for filtering the search terms, this task is often difficult because finding related keywords could be hard for some diseases or be costly when conducting for multiple diseases. The second is the selection of proper models. While the literature on data-driven nowcasting methods for estimating infectious disease activity is well-developed from an epidemiological standpoint, the machine learning methods employed lag behind the state-of-the-art. The nowcasting models introduced to date mainly use variations of regularized linear regressions or, less often, random forests or support vector machines. From a machine learning perspective, the problem of disease activity estimation is most suited to more sophisticated and time-series specific model architecture, and thanks to the growing volume of recorded environmental-mediated disease data, the use of recurrent neural networks (RNNs), and more specifically their variants long short-term memory (LSTM) and gated recurrent unit (GRU) networks, is increasingly feasible. The vanilla LSTM model makes predictions solely relying on the time series of search activity while ignoring the semantic information in the search query phrases. Previous studies have pointed out that the search queries could be semantically related and ignoring their correlation would lead to a decrease in the model performance [28, 33]. Recent advances in natural language processing (NLP) have led to the development of a technique called word embeddings to represent the semantic information in the phrases and fine-tuning of word embeddings has been encouraged for down-stream tasks [52]. However, there is still a lack of knowledge of incorporating both the semantic information of search queries and the time series of search activity to make predictions. 

\vspace{1\baselineskip}
In this study, we investigate Web\textit{ }search\textit{ }data as one important source of an online crowd-based indicator. As Web search data is free and broadly accessible, we posit that it could serve as a scalable means of tracking urban air pollution exposures and corresponding population-level health responses. To measure search interest, we use the freely accessible Google Trends service, which reports aggregate search volume data at city-level geographical resolution. For this analysis we use known health endpoint terms and topics, such as ``difficulty breathing", and observations (e.g., ``haze") suggested by public health researchers, augmented by automatic term expansion based on semantic and temporal correlations, to estimate the levels of search activities related to air pollution, and ultimately to predict whether the pollution levels were elevated. 

\vspace{1\baselineskip}
Compared to existing air pollution classification models, this study explores the usage of Web search anomalies as an auxiliary signal to detect air pollution. We compare our approach to the state-of-the-art physical sensor-based models which incorporate various pollutant covariates such as historical pollutant concentrations and meteorological data [19]. Using Web search data for prediction introduces several challenges, including an unclear relationship between search interest and pollution levels and the trade-off between model complexity and convergence for the inclusion of Web search data on a data-deficient scenario.

\vspace{1\baselineskip}
In summary, our contributions are:
\begin{itemize}
	\item \begin{flushleft}
{ We propose a novel search term Dictionary Learner-Long-Short Term Memory (DL-LSTM) model to learn sequential patterns from broad historical records of Web search data for air pollution nowcasting.}
\end{flushleft}
\vspace*{-2.5cm}
	\item \begin{flushleft}
{ We compare the DL-LSTM models to a variety of baseline models on the efficacy of using Web search data to indicate exposure to a non-infectious environment stressor (i.e., air pollution) and demonstrate that the proposed models are effective across different experimental settings. }
\end{flushleft}
\vspace*{-2.5cm}
	\item \begin{flushleft}
{ We evaluate the efficacy of combining Web search data and meteorology data for air pollution predicting and show that the inclusion of Web search data improves the prediction accuracy and provides a promising substitute when historical pollutant data is unavailable.  }
\end{flushleft}
\end{itemize}

\begin{table}[H]
\begin{adjustbox}{max width=\textwidth}
\begin{tabular}{p{7.41cm}p{7.83cm}p{7.41cm}p{7.83cm}}
\hline
\multicolumn{1}{|p{7.41cm}}{Input Feature} & 
\multicolumn{1}{|p{7.83cm}|}{Feature Transformation} \\ 
\hline
\multicolumn{1}{|p{7.41cm}}{} & 
\multicolumn{1}{|p{7.83cm}|}{} \\ 
\hline
\multicolumn{1}{|p{7.41cm}}{\textbf{Meteorological Data (Met)}} & 
\multicolumn{1}{|p{7.83cm}|}{} \\ 
\hline
\multicolumn{1}{|p{7.41cm}}{} & 
\multicolumn{1}{|p{7.83cm}|}{Max Temperature (Temp\_max)} \\ 
\hline
\multicolumn{1}{|p{7.41cm}}{} & 
\multicolumn{1}{|p{7.83cm}|}{Mean Temperature (Temp\_mean)} \\ 
\hline
\multicolumn{1}{|p{7.41cm}}{} & 
\multicolumn{1}{|p{7.83cm}|}{Relative Humidity (Humidity)} \\ 
\hline
\multicolumn{1}{|p{7.41cm}}{} & 
\multicolumn{1}{|p{7.83cm}|}{Square of Temp\_mean} \\ 
\hline
\multicolumn{1}{|p{7.41cm}}{} & 
\multicolumn{1}{|p{7.83cm}|}{Cube of Temp\_mean} \\ 
\hline
\multicolumn{1}{|p{7.41cm}}{} & 
\multicolumn{1}{|p{7.83cm}|}{Square of Humidity} \\ 
\hline
\multicolumn{1}{|p{7.41cm}}{} & 
\multicolumn{1}{|p{7.83cm}|}{Cube of Humidity} \\ 
\hline
\multicolumn{1}{|p{7.41cm}}{} & 
\multicolumn{1}{|p{7.83cm}|}{Dew Point Temperature} \\ 
\hline
\multicolumn{1}{|p{7.41cm}}{\textbf{Pollutant Concentration (Pol)}} & 
\multicolumn{1}{|p{7.83cm}|}{} \\ 
\hline
\multicolumn{1}{|p{7.41cm}}{} & 
\multicolumn{1}{|p{7.83cm}|}{Concentration on day t-7} \\ 
\hline
\multicolumn{1}{|p{7.41cm}}{} & 
\multicolumn{1}{|p{7.83cm}|}{Concentration on day t-6} \\ 
\hline
\multicolumn{1}{|p{7.41cm}}{} & 
\multicolumn{1}{|p{7.83cm}|}{\centering
$\ldots$} \\ 
\hline
\multicolumn{1}{|p{7.41cm}}{} & 
\multicolumn{1}{|p{7.83cm}|}{Concentration on day t-1} \\ 
\hline
\multicolumn{1}{|p{7.41cm}}{\textbf{Search}} & 
\multicolumn{1}{|p{7.83cm}|}{} \\ 
\hline
\multicolumn{1}{|p{7.41cm}}{} & 
\multicolumn{1}{|p{7.83cm}|}{Search volumes of search terms } \\ 
\hline
\end{tabular}
\end{adjustbox}
\caption{Input features calculated per time step in the input sequence.}
\end{table}

\subsection*{Methods}
We now describe the methodology: first, we formalize our problem setting, then describe the data and then introduce our modeling approaches.

\subsubsection*{Problem Statement}

We formalized this task as a classification problem and adapted state-of-the-art machine learning models. We constructed a multivariate autoregressive model and a random forest model fit on historical air pollutant concentration as well as search and meteorological data as baseline models. We evaluate the performance our proposed models (described below) compared to the baselines on prediction accuracy and other standard classification prediction metrics.

\subsubsection*{Data Collection}

We collected daily air pollutant concentration data as well as temperature and relative humidity in the ten largest U.S. MSAs from Jan. 2007 to Dec. 2018. We focus on three air pollutants: O\textsubscript{3}, NO\textsubscript{2}, and PM\textsubscript{2.5}. The in-situ pollutant concentrations and meteorological data such as temperature, relative humidity, and dew-point temperature were retrieved from the US Environmental Protection Agency (EPA), Air Quality System (AQS), and AirNow database. To create a single daily pollutant concentration for each city, we used the median pollutant concentration from all available monitoring sites within each city to avoid outlier bias.

We collected the daily search frequency of pollution-related terms from Google Trends for the same 12-year period and cities. We created a curated list of 152 pollution-related terms based on our previous air pollution epidemiology studies and in reviewing the environmental health literature [41-43, 57-59] and downloaded the reports of trending results terms using PyTrends [44]. For each PyTrends request, we downloaded the search history of pollution-related terms over a six-month window with one overlapping month for calibration. PyTrends provided us with search frequency scaled on a range of 0 to 100 based on a topic’s proportion to all searches on all topics. Because of the PyTrends restriction, we downloaded the reports of trending results multiple times and the search frequencies are scaled, separately in each six-month window, which required us to calibrate the search frequency for the 12-year period. We calibrateed the search frequencies by joining the search logs on the overlapping periods (1 out of 6 months) for inter-calibration [45].

\vspace{1\baselineskip}
We investigated the available input features from meteorological data (temperature and relative humidity), historical pollutant concentration, and Web search data (see (Table 1)).

\subsubsection*{Missing Data Imputation and Normalization}

Smoothing and interpolation is a simple and efficient data imputation method [53],

we apply linear interpolation to fill the missing data in historical pollutant concentration, temperature, and humidity, with a rolling mean of window size 3. To fill in the missing data in infrequent search terms for which Google Trends does not return a count, we use random numbers close to zero (\( e^{-10}\sim e^{-5}\)). We normalize all the input features to standard scores by subtracting their mean values and dividing them by the respective standard deviations. 

\subsubsection*{Search Term Expansion (STE)}

As online search queries may reflect individual exposures to ambient air pollution, the seed terms are mostly related to symptoms, observations, and emission sources (Appendix 1; Table S1). However, since an exhaustive list of user queries was not available, reliance on only expert-generated seed words may result in a poor prediction--due to the high mismatch rate between the user queries and our expected search words. 

\vspace{1\baselineskip}
Query expansion is a common approach to resolve this discrepancy. A recent study [28] showed that the initial set of seed words can be effectively expanded through semantic and temporal correlations. Thus, for each seed word we use Google Correlate [46] to retrieve the top 100 correlated query terms. Then we use the pre-trained word2vec model [47] to retrieve the vector representation of each query--phrases are mapped to the centroid of the constituent terms. A utility score is calculated for each candidate query by measuring the maximum cosine similarity between the query and the seed words. The queries with a high utility score are retained and the remaining queries are eliminated--we empirically set the utility cut-off to 0.55. This method expands the set of search terms for a total of 152 search terms to track (Appendix 1; Table S2).

\subsubsection*{Modeling and Evaluation}

\paragraph{Problem Definition}

Given sequences of physical sensor data \( P =\left[p_{t-L}, \ldots ,p_{t-1}\right]^{T}\in R^{L\times d_{p}}\), and search interest data \( S =\left[s_{t-L+2}, \ldots , s_{t+1}\right]^{T}\in R^{L\times d_{s}}\), the task is to classify day \( t\) as ``polluted" or not, where the positive class label indicates that the air pollution was above a pre-defined threshold. \( L\) denotes sequence length, and \( d_{p}\) and \( d_{s}\) are the number of physical sensors features and the number of search-related terms, respectively.

\paragraph{Autoregressive and Random Forest Classification Models}

Previous work has shown that simple autoregressive models using Web search data can generate nowcast estimates for influenza-like illness (ILI) at the US national level [33]. We adapt the autoregressive models with the logistic regression (LR) classifier for classification purposes. Furthermore, we apply elastic net regularization, which is a linear combination of \( l_{1}\) and \( l_{2}\) regularization and is proposed in previous studies [28, 33]. LR+Elastic Net was implemented using python \textit{scikit-learn} package, using cross-validation to set the model's hyper-parameters to maximize the F1 score on the validation set, with class\_weight set to "balanced".

\vspace{1\baselineskip}
Random forest is an ensemble learning model and is robust against over-fitting, providing a strong baseline for the development of nonlinear predictive models [50]. We used the \textit{scikit-learn} implementation of random forests. The number of trees and the maximum depth of individual trees were selected to maximize the F1 score on the validation set, with balanced class\_weight for positive and negative samples.

\vspace{1\baselineskip}

\vspace{1\baselineskip}
\begin{figure}[H]
\centering
\includegraphics[width=12.0cm,height=7.44cm]{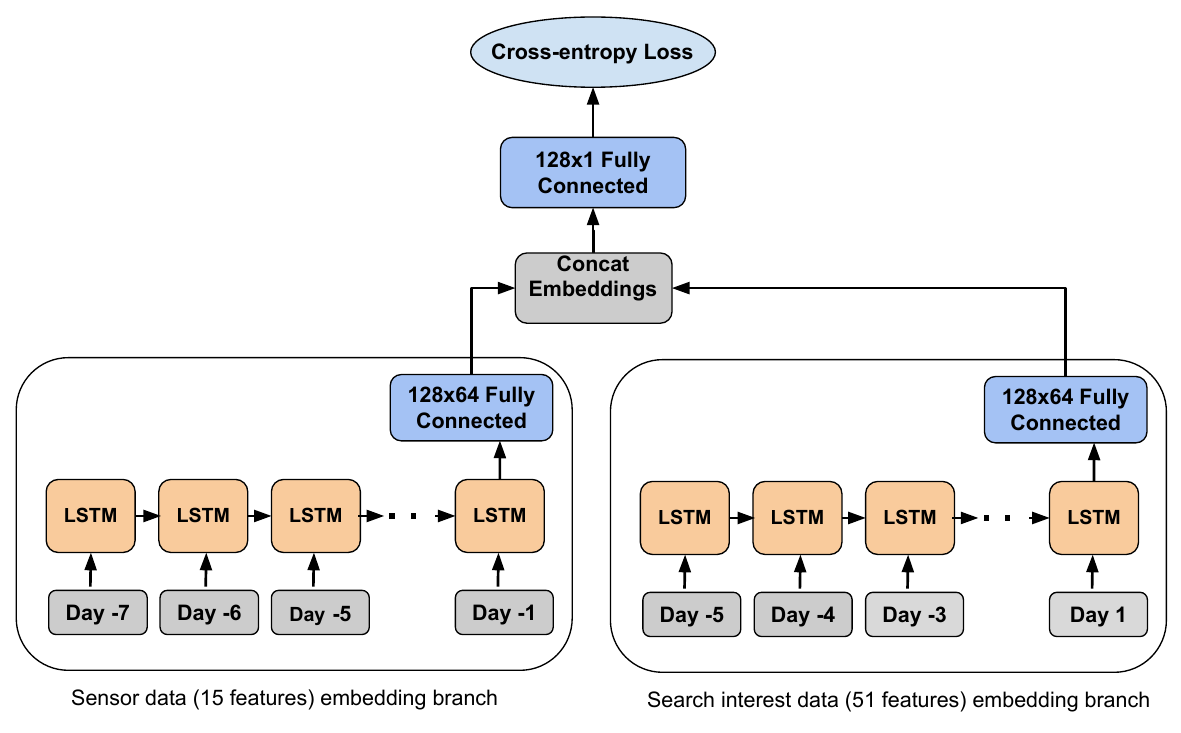}
\caption{The Architecture of the LSTM Model.}
\end{figure}

\begin{figure}[H]
\centering
\includegraphics[width=5.86cm,height=7.44cm]{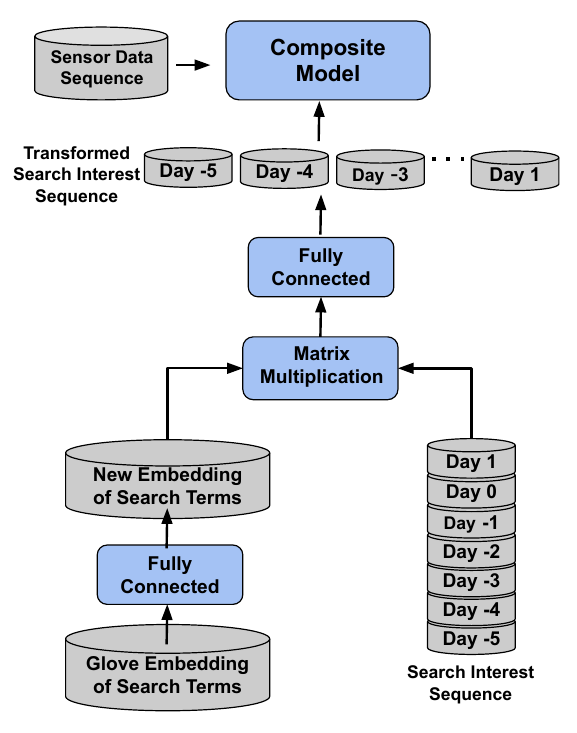}
\caption{The Architecture of the DL-LSTM Model.}
\end{figure}




\paragraph{Long Short-Term Memory and its Variants}

Long short-term memory units (LSTM) [37] are recurrent neural networks (RNNs) models designed for sequence modeling, which could learn the non-linear relationship in time series data [38]. We first describe a baseline LSTM model with two sub-networks to separate the search data and meteorological data. As shown in (Figure 1), there are four layers in the model, i.e., the sequence embedding layer, the LSTM layer, the fully-connected hidden layer, and the output layer. 

\vspace{1\baselineskip}
In the left sub-network of the LSTM model with search data as input, we propose two methods to capture the semantic information in search terms. The first one is called the LSTM semantic model (LSTM-GloVe). As a variant of the vanilla LSTM model, for the sequence embedding layer of the right sub-network in (Figure 1), we introduce the matrix multiplication operation to project the search values of search terms to their semantic embedding space (GloVe embeddings) as shown in (Equation 1). 

\vspace{1\baselineskip}
Given the search interest data \( S = [s_{1}, \ldots , s_{7}]^{T}\in R^{7\times d_{s}}\), and their GloVe embedding 

\( G = [g_{1}, \ldots ,g_{d_{g}}]\in R^{d_{s}\times d_{g}}\), where \( d_{g} = 50\) (GloVe 50-dimensional word vectors trained on tweets [52]). The matrix multiplication operation is defined as:
\begin{equation}
S \cdot G=\left[\begin{array}{ccc}
- & s_1 & - \\
- & s_2 & - \\
& \vdots & \\
- & s_7 & -
\end{array}\right] \cdot\left[g_1 g_2 \ldots g_{d_g}\right] \in R^{7 \times d_g}
\end{equation}


\vspace{1\baselineskip}
Specifically, the tensor generated by matrix multiplication operation is then fed to LSTM layer for further calculation. This matrix multiplication is designed specifically for the model consistency problem when introducing co-linear predictors after search term expansion.

\vspace{1\baselineskip}
The second variation of the LSTM model is the DL-LSTM model, based theoretically on the idea of matrix multiplication as shown in LSTM-GloVe. However, instead of directly applying the GloVe embedding for matrix multiplication, it introduces the fine-tuning of the word embeddings via a \( d_{g}\) by \( d_{e}\) ReLU-activated fully-connected layer. As shown in (Figure 2), it applies the ReLU-activated fully-connected layer to the initial GloVe embedding, where \( d_{e} = 100\) is the size of the new embedding. In this architecture, the GloVe 50-dimensional word vectors are used to initialize the search term embedding dictionary, and the matrix multiplication operation is used to transform the input embedding of search terms to the semantic embedding space\footnote{https://github.com/emory-irlab/airpollutionnowcast}.

\vspace{1\baselineskip}
In summary, we evaluate the following models in this paper:

\begin{itemize}
	\item \begin{flushleft}
{\large \textbf{\textit{LR}}: Logistic regression classifier with elastic net regularization.}
\end{flushleft}

	\item \begin{flushleft}
{\large \textbf{\textit{RF}}: Random forest classifier with the number of trees and maximum depth tuned for prediction.}
\end{flushleft}

	\item \begin{flushleft}
{\large \textbf{\textit{LSTM}}: Baseline LSTM model as shown in (Figure 1), which combines physical sensor features, if available, with the search interest volume data directly, providing a direct adaptation of RNNs to this problem without any problem-specific extensions.\par}
\end{flushleft}

	\item \begin{flushleft}
{\large \textit{\textbf{LSTM-GloVe}}: LSTM semantic model, which is a variant of LSTM model as described by (Equation 1), we control the input of search interest data (\textit{i.e.}, 51 seed search terms vs. 152 terms after STE) in this model. We refer to the variants as \textit{LSTM-GloVe} and \textit{LSTM-GloVe w/ STE} respectively.\par}
\end{flushleft}

	\item \begin{flushleft}
{\large \textit{\textbf{DL-LSTM}}: DL-LSTM model as shown in (Figure 2). We control the input of search interest data (i.e. 51 seed search terms vs. 152 terms after STE) in this model and refer to the variants as \textit{DL-LSTM} and \textit{DL-LSTM w/ STE}\textbf{ }respectively.\par}
\end{flushleft}

\end{itemize}

\vspace{1\baselineskip}

\begin{table}[H]
\begin{adjustbox}{max width=\textwidth}
\begin{tabular}{p{2.65cm}p{2.0cm}p{2.83cm}p{1.73cm}p{1.72cm}p{2.83cm}p{1.46cm}p{2.65cm}p{2.0cm}p{2.83cm}p{1.73cm}p{1.72cm}p{2.83cm}p{1.46cm}}
\hline
\multicolumn{1}{|p{2.65cm}}{\multirow{2}{*}{\parbox{2.65cm}{Pollutant}}} & 
\multicolumn{3}{|p{6.5600000000000005cm}}{Negative Samples} & 
\multicolumn{3}{|p{6.01cm}|}{Positive Samples} \\ 
\hhline{~------}
\multicolumn{1}{|p{2.65cm}}{} & 
\multicolumn{1}{|p{2.0cm}}{train} & 
\multicolumn{1}{|p{2.83cm}}{validation} & 
\multicolumn{1}{|p{1.73cm}}{test} & 
\multicolumn{1}{|p{1.72cm}}{train} & 
\multicolumn{1}{|p{2.83cm}}{validation} & 
\multicolumn{1}{|p{1.46cm}|}{test} \\ 
\hline
\multicolumn{1}{|p{2.65cm}}{} & 
\multicolumn{1}{|p{2.0cm}}{} & 
\multicolumn{1}{|p{2.83cm}}{} & 
\multicolumn{1}{|p{1.73cm}}{} & 
\multicolumn{1}{|p{1.72cm}}{} & 
\multicolumn{1}{|p{2.83cm}}{} & 
\multicolumn{1}{|p{1.46cm}|}{} \\ 
\hline
\multicolumn{1}{|p{2.65cm}}{\textbf{O\textsubscript{3}}} & 
\multicolumn{1}{|p{2.0cm}}{} & 
\multicolumn{1}{|p{2.83cm}}{} & 
\multicolumn{1}{|p{1.73cm}}{} & 
\multicolumn{1}{|p{1.72cm}}{} & 
\multicolumn{1}{|p{2.83cm}}{} & 
\multicolumn{1}{|p{1.46cm}|}{} \\ 
\hline
\multicolumn{1}{|p{2.65cm}}{} & 
\multicolumn{1}{|p{2.0cm}}{\raggedleft
24322} & 
\multicolumn{1}{|p{2.83cm}}{\raggedleft
6269} & 
\multicolumn{1}{|p{1.73cm}}{\raggedleft
6311} & 
\multicolumn{1}{|p{1.72cm}}{\raggedleft
4896} & 
\multicolumn{1}{|p{2.83cm}}{\raggedleft
1038} & 
\multicolumn{1}{|p{1.46cm}|}{\raggedleft
982} \\ 
\hline
\multicolumn{1}{|p{2.65cm}}{\textbf{NO\textsubscript{2}}} & 
\multicolumn{1}{|p{2.0cm}}{} & 
\multicolumn{1}{|p{2.83cm}}{} & 
\multicolumn{1}{|p{1.73cm}}{} & 
\multicolumn{1}{|p{1.72cm}}{} & 
\multicolumn{1}{|p{2.83cm}}{} & 
\multicolumn{1}{|p{1.46cm}|}{} \\ 
\hline
\multicolumn{1}{|p{2.65cm}}{} & 
\multicolumn{1}{|p{2.0cm}}{\raggedleft
23926} & 
\multicolumn{1}{|p{2.83cm}}{\raggedleft
6119} & 
\multicolumn{1}{|p{1.73cm}}{\raggedleft
6332} & 
\multicolumn{1}{|p{1.72cm}}{\raggedleft
5292} & 
\multicolumn{1}{|p{2.83cm}}{\raggedleft
1188} & 
\multicolumn{1}{|p{1.46cm}|}{\raggedleft
961} \\ 
\hline
\multicolumn{1}{|p{2.65cm}}{\textbf{PM\textsubscript{2.5}}} & 
\multicolumn{1}{|p{2.0cm}}{} & 
\multicolumn{1}{|p{2.83cm}}{} & 
\multicolumn{1}{|p{1.73cm}}{} & 
\multicolumn{1}{|p{1.72cm}}{} & 
\multicolumn{1}{|p{2.83cm}}{} & 
\multicolumn{1}{|p{1.46cm}|}{} \\ 
\hline
\multicolumn{1}{|p{2.65cm}}{} & 
\multicolumn{1}{|p{2.0cm}}{\raggedleft
24297} & 
\multicolumn{1}{|p{2.83cm}}{\raggedleft
6745} & 
\multicolumn{1}{|p{1.73cm}}{\raggedleft
6757} & 
\multicolumn{1}{|p{1.72cm}}{\raggedleft
4921} & 
\multicolumn{1}{|p{2.83cm}}{\raggedleft
562} & 
\multicolumn{1}{|p{1.46cm}|}{\raggedleft
536} \\ 
\hline
\end{tabular}
\end{adjustbox}
\caption{The distribution of classes in train, validation and test sets.}
\end{table}

\subsubsection*{Validation}

To tune model parameters and validate model performance, we split the available data into training (from Jan. 2007 to Dec. 2014), validation (from Jan. 2015 to Dec. 2016), and testing (from Jan. 2017 to Dec. 2018) sets. This eight-year training period provides a broad history to learn a relationship between input predictors and output variables, and the predictive models are evaluated based on their ability to make predictions for completely unseen periods. For evaluating our model, we make predictions for each day form Jan. 2017 to Dec. 2018 in the test dataset. The distribution of classes in train, validation, and test datasets is reported in (Table 2). Note that positive and negative classes are heavily imbalanced, with positive classes comprising, for instance, only 16$\%$ of training samples when PM\textsubscript{2.5 }is the target pollutant.

\subsubsection*{Evaluation metrics}

Because we defined this task as a classification problem, we used standard classification evaluation metrics. We report accuracy and F1 score of the positive class (the harmonic mean of precision and recall) of predictions as evaluation metrics for all models. While accuracy measures the total fraction of correct predictions and could misrepresent model performance in presence of heavily imbalanced classes, the F1 score takes class imbalance into account and is, therefore, a more appropriate metric for our problem.

\begin{table}[H]
\begin{adjustbox}{max width=\textwidth}
\begin{tabular}{p{1.07cm}p{13.11cm}p{1.07cm}p{1.07cm}p{13.11cm}p{1.07cm}}
\multicolumn{1}{p{1.07cm}}{} & 
\multicolumn{1}{p{13.11cm}}{\( Accuracy = \frac{TP+TN}{TP+TN+FP+FN}\)} & 
\multicolumn{1}{p{1.07cm}}{\raggedleft
(2)} \\ 
\multicolumn{1}{p{1.07cm}}{} & 
\multicolumn{1}{p{13.11cm}}{\( F1 = \frac{2}{\frac{1}{recall}+\frac{1}{precision} }\)} & 
\multicolumn{1}{p{1.07cm}}{\raggedleft
(3)} \\ 
\end{tabular}
\end{adjustbox}
\end{table}
where \( TP\), \( TN\), \( FP\), and \( FN\) are the number of true positive samples, true negative samples, false positive samples, and false negative samples, respectively.

\subsection*{Results}
\subsubsection*{Overview}

In this section, we first present the findings of the data exploration. Then we present the principal findings of this study.

\subsubsection*{Insights from Collected Data}

In this section, we describe the thresholds of abnormal air pollutant concentrations and then we present the lag between search anomalies and air pollution.  

\paragraph{Thresholds of Abnormal Air pollutant Concentrations}

The major MSAs chosen for study in this work, have different distributions of pollutant concentration through time, almost always fall below the EPA standard 24-hour threshold (Figure 3). Despite this, multiple studies have shown that even at low concentrations, chronic exposure to air pollution negatively affects human health [41, 42]. Therefore, calibrating a meaningful threshold for each city, especially ones with generally lower levels of air pollution (e.g., Miami) may be critical for adequately protecting population health. A natural way to do this may be to set the threshold as one standard deviation above the mean daily pollutant concentration within each city, which is adopted in this study. The input predictors are also normalized within each city to reflect city-level dynamics. The resulting thresholds for the three pollutants and cities under investigation are reported in (Table 3).


\paragraph{Lag between Search Anomalies and Air Pollution}

A previous study showed that there could be a lag between incident occurrence and google search activity. [31]. As shown in (Figure 4), the normalized search frequency of the term ``cough" is correlated with the concentration of NO\textsubscript{2} in Atlanta with a certain lag of time. To determine the lag between elevated pollution levels and consequent pollution-related searches, the mean absolute Spearman's correlation between pollutant concentrations and search interest data was calculated, shifted forward in time for 0, 1, 2, and 3 days. As shown in (Table 4), for O\textsubscript{3} and PM\textsubscript{2.5}, the mean absolute Spearman's correlation increases with the increase of the shifted days. Considering that the task aims to detect elevated pollution levels as soon as possible, a lag of one day was applied to search data. In other words, search interest data from the current day was used to estimate whether air pollution was elevated on the previous day.

\subsubsection*{Evaluation Outcomes}

In this section, we consider three conditions to evaluate the performance of using Web search data to detect elevated pollution, i.e., using only search data, using search data as an auxiliary data of meteorological data, and using search data as an auxiliary data of meteorological data and historical pollutant concentration.
\paragraph{Using Only Search Data:}

For areas where ambient pollution monitoring is unavailable, investigating whether Web search data could be used as the only signal for nowcasting elevated air pollution is a vital question. When relying on only search data for air pollution prediction, both the proposed DL-LSTM architecture and search term expansion contribute to the improvement of prediction accuracy. As shown in the ``Search" section of (Table 5), the LSTM-based models exhibit superior accuracy over the baseline LR and RF models for O\textsubscript{3 }and NO\textsubscript{2.} For PM\textsubscript{2.5}, the proposed models do not perform better than the baseline LR or LSTM model because the validation and test dataset are heavily imbalanced (as shown in (Table 5)). In more detail, the proposed DL-LSTM w/ STE model achieves the highest F1 score (32.44$\%$ for O\textsubscript{3}, 27.70$\%$ for NO\textsubscript{2}) for detecting O\textsubscript{3 }and NO\textsubscript{2} pollution.

\begin{figure}[h]
\includegraphics[width=12.0cm,height=15.53cm]{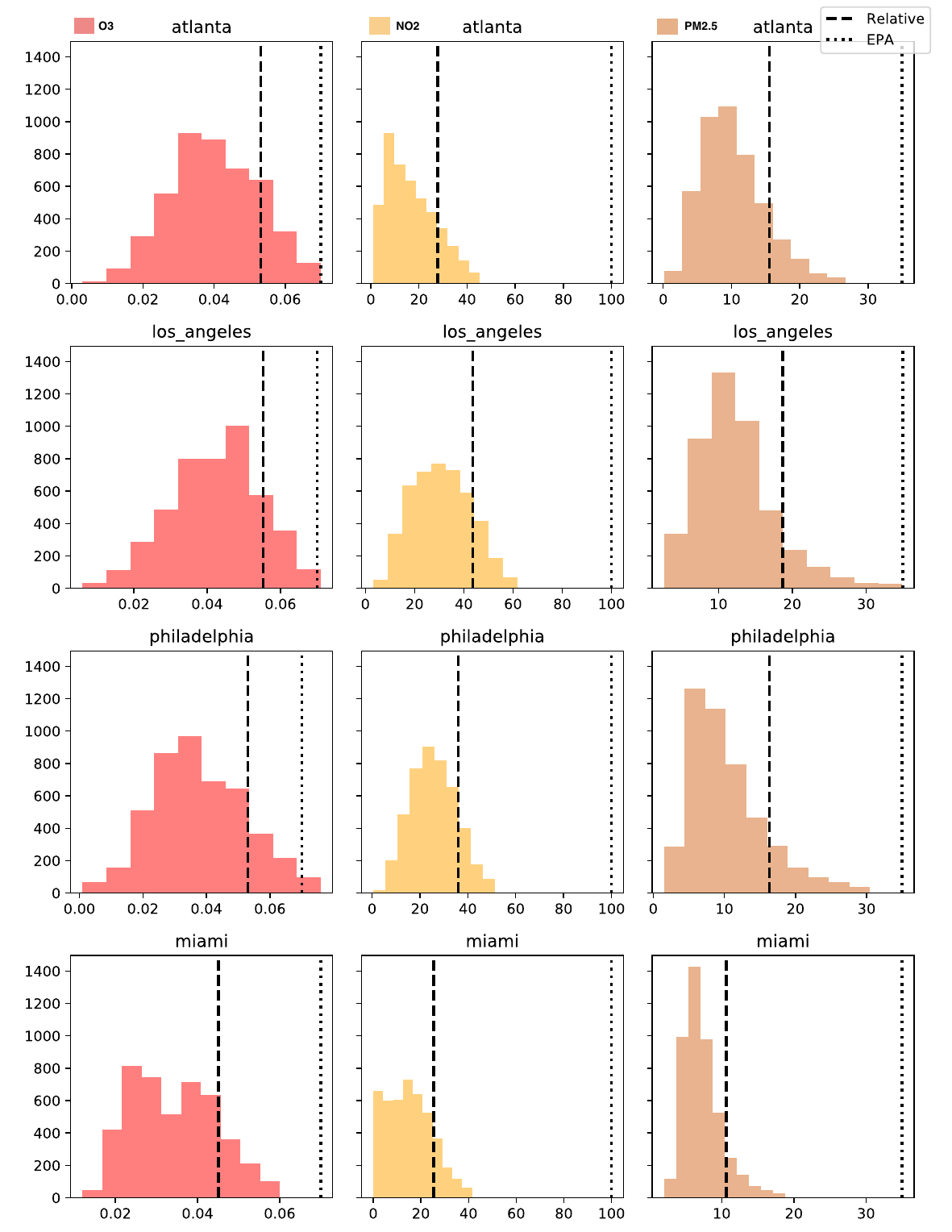}
\caption{Distribution of pollution values for Atlanta, Los Angeles, Philadelphia, and Miami, with city-specific elevated pollution level (dashed line) and the general EPA-mandated standard (dotted line), for O\textsubscript{3} (left column), NO\textsubscript{2} (middle column), and PM\textsubscript{2.5}(right column).}
\end{figure}


\begin{table}[h]
\begin{adjustbox}{max width=\textwidth}
\begin{tabular}{p{2.39cm}p{1.23cm}p{1.23cm}p{1.44cm}p{1.6cm}p{1.23cm}p{1.23cm}p{1.23cm}p{1.23cm}p{1.23cm}p{1.23cm}p{2.39cm}p{1.23cm}p{1.23cm}p{1.44cm}p{1.6cm}p{1.23cm}p{1.23cm}p{1.23cm}p{1.23cm}p{1.23cm}p{1.23cm}}
\hline
\multicolumn{1}{|p{2.39cm}}{Pollutant } & 
\multicolumn{1}{|p{1.23cm}}{ L.A.} & 
\multicolumn{1}{|p{1.23cm}}{ DC } & 
\multicolumn{1}{|p{1.44cm}}{ PHILA. } & 
\multicolumn{1}{|p{1.6cm}}{ DTX} & 
\multicolumn{1}{|p{1.23cm}}{ ATL } & 
\multicolumn{1}{|p{1.23cm}}{BOS } & 
\multicolumn{1}{|p{1.23cm}}{ NY } & 
\multicolumn{1}{|p{1.23cm}}{ MIA } & 
\multicolumn{1}{|p{1.23cm}}{ CHI } & 
\multicolumn{1}{|p{1.23cm}|}{ HOU} \\ 
\hline
\multicolumn{1}{|p{2.39cm}}{} & 
\multicolumn{1}{|p{1.23cm}}{} & 
\multicolumn{1}{|p{1.23cm}}{} & 
\multicolumn{1}{|p{1.44cm}}{} & 
\multicolumn{1}{|p{1.6cm}}{} & 
\multicolumn{1}{|p{1.23cm}}{} & 
\multicolumn{1}{|p{1.23cm}}{} & 
\multicolumn{1}{|p{1.23cm}}{} & 
\multicolumn{1}{|p{1.23cm}}{} & 
\multicolumn{1}{|p{1.23cm}}{} & 
\multicolumn{1}{|p{1.23cm}|}{} \\ 
\hline
\multicolumn{1}{|p{2.39cm}}{\textbf{O\textsubscript{3}(ppb)}} & 
\multicolumn{1}{|p{1.23cm}}{} & 
\multicolumn{1}{|p{1.23cm}}{} & 
\multicolumn{1}{|p{1.44cm}}{} & 
\multicolumn{1}{|p{1.6cm}}{} & 
\multicolumn{1}{|p{1.23cm}}{} & 
\multicolumn{1}{|p{1.23cm}}{} & 
\multicolumn{1}{|p{1.23cm}}{} & 
\multicolumn{1}{|p{1.23cm}}{} & 
\multicolumn{1}{|p{1.23cm}}{} & 
\multicolumn{1}{|p{1.23cm}|}{} \\ 
\hline
\multicolumn{1}{|p{2.39cm}}{} & 
\multicolumn{1}{|p{1.23cm}}{\raggedleft
55} & 
\multicolumn{1}{|p{1.23cm}}{\raggedleft
54} & 
\multicolumn{1}{|p{1.44cm}}{\raggedleft
53} & 
\multicolumn{1}{|p{1.6cm}}{\raggedleft
53} & 
\multicolumn{1}{|p{1.23cm}}{\raggedleft
53} & 
\multicolumn{1}{|p{1.23cm}}{\raggedleft
48} & 
\multicolumn{1}{|p{1.23cm}}{\raggedleft
49} & 
\multicolumn{1}{|p{1.23cm}}{\raggedleft
45} & 
\multicolumn{1}{|p{1.23cm}}{\raggedleft
49} & 
\multicolumn{1}{|p{1.23cm}|}{\raggedleft
49} \\ 
\hline
\multicolumn{1}{|p{2.39cm}}{\textbf{NO\textsubscript{2}(ppb)}} & 
\multicolumn{1}{|p{1.23cm}}{} & 
\multicolumn{1}{|p{1.23cm}}{} & 
\multicolumn{1}{|p{1.44cm}}{} & 
\multicolumn{1}{|p{1.6cm}}{} & 
\multicolumn{1}{|p{1.23cm}}{} & 
\multicolumn{1}{|p{1.23cm}}{} & 
\multicolumn{1}{|p{1.23cm}}{} & 
\multicolumn{1}{|p{1.23cm}}{} & 
\multicolumn{1}{|p{1.23cm}}{} & 
\multicolumn{1}{|p{1.23cm}|}{} \\ 
\hline
\multicolumn{1}{|p{2.39cm}}{} & 
\multicolumn{1}{|p{1.23cm}}{\raggedleft
43.7} & 
\multicolumn{1}{|p{1.23cm}}{\raggedleft
38.1} & 
\multicolumn{1}{|p{1.44cm}}{\raggedleft
36} & 
\multicolumn{1}{|p{1.6cm}}{\raggedleft
25.2} & 
\multicolumn{1}{|p{1.23cm}}{\raggedleft
27.8} & 
\multicolumn{1}{|p{1.23cm}}{\raggedleft
30.7} & 
\multicolumn{1}{|p{1.23cm}}{\raggedleft
45.3} & 
\multicolumn{1}{|p{1.23cm}}{\raggedleft
25.5} & 
\multicolumn{1}{|p{1.23cm}}{\raggedleft
43.7} & 
\multicolumn{1}{|p{1.23cm}|}{\raggedleft
27.7} \\ 
\hline
\multicolumn{1}{|p{2.39cm}}{\textbf{PM\textsubscript{2.5}(ug/m\textsuperscript{3})}} & 
\multicolumn{1}{|p{1.23cm}}{} & 
\multicolumn{1}{|p{1.23cm}}{} & 
\multicolumn{1}{|p{1.44cm}}{} & 
\multicolumn{1}{|p{1.6cm}}{} & 
\multicolumn{1}{|p{1.23cm}}{} & 
\multicolumn{1}{|p{1.23cm}}{} & 
\multicolumn{1}{|p{1.23cm}}{} & 
\multicolumn{1}{|p{1.23cm}}{} & 
\multicolumn{1}{|p{1.23cm}}{} & 
\multicolumn{1}{|p{1.23cm}|}{} \\ 
\hline
\multicolumn{1}{|p{2.39cm}}{} & 
\multicolumn{1}{|p{1.23cm}}{\raggedleft
18.7} & 
\multicolumn{1}{|p{1.23cm}}{\raggedleft
15.1} & 
\multicolumn{1}{|p{1.44cm}}{\raggedleft
16.4} & 
\multicolumn{1}{|p{1.6cm}}{\raggedleft
13.1} & 
\multicolumn{1}{|p{1.23cm}}{\raggedleft
15.6} & 
\multicolumn{1}{|p{1.23cm}}{\raggedleft
12.4} & 
\multicolumn{1}{|p{1.23cm}}{\raggedleft
13.9} & 
\multicolumn{1}{|p{1.23cm}}{\raggedleft
10.6} & 
\multicolumn{1}{|p{1.23cm}}{\raggedleft
16.2} & 
\multicolumn{1}{|p{1.23cm}|}{\raggedleft
14.4} \\ 
\hline
\end{tabular}
\end{adjustbox}
\caption{Classification thresholds for three pollutants across 10 major MSAs in the U.S.}
\end{table}



\begin{figure}[h]
\includegraphics[width=12.0cm,height=5.06cm]{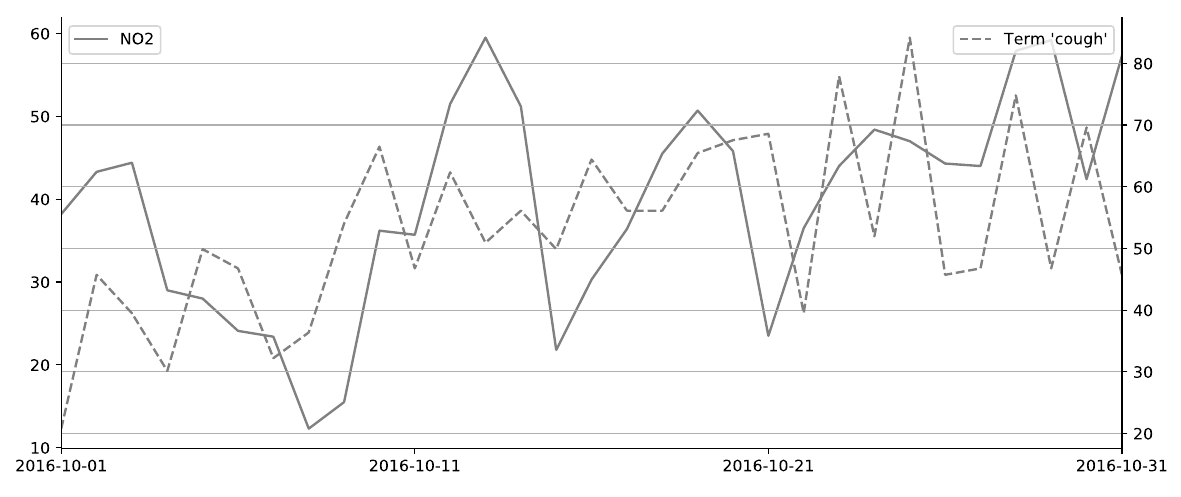}
\caption{NO\textsubscript{2} levels and search interest for term "cough" in Atlanta, October 2016.}
\end{figure}


\begin{table}[h]
\begin{adjustbox}{max width=\textwidth}
\begin{tabular}{p{1.66cm}p{2.85cm}p{1.25cm}p{2.86cm}p{1.2cm}p{3.19cm}p{1.5cm}p{2.91cm}p{1.52cm}p{1.66cm}p{2.85cm}p{1.25cm}p{2.86cm}p{1.2cm}p{3.19cm}p{1.5cm}p{2.91cm}p{1.52cm}}
\hline
\multicolumn{1}{|p{1.66cm}}{\centering
Pollutant} & 
\multicolumn{1}{|p{2.85cm}}{Lag $=$ 0 \newline
Search Term (Spearman's correlation)} & 
\multicolumn{1}{|p{1.25cm}}{\textit{P\textsuperscript{a}}} & 
\multicolumn{1}{|p{2.86cm}}{Lag$=$1 \newline
Search Term (Spearman's correlation)} & 
\multicolumn{1}{|p{1.2cm}}{\textit{P\textsuperscript{a}}} & 
\multicolumn{1}{|p{3.19cm}}{Lag$=$2 \newline
Search Term (Spearman's correlation)} & 
\multicolumn{1}{|p{1.5cm}}{\textit{P\textsuperscript{a}}} & 
\multicolumn{1}{|p{2.91cm}}{Lag$=$3 \newline
Search Term (Spearman's correlation)} & 
\multicolumn{1}{|p{1.52cm}|}{\textit{P\textsuperscript{a}}} \\ 
\hline
\multicolumn{1}{|p{1.66cm}}{} & 
\multicolumn{1}{|p{2.85cm}}{} & 
\multicolumn{1}{|p{1.25cm}}{} & 
\multicolumn{1}{|p{2.86cm}}{} & 
\multicolumn{1}{|p{1.2cm}}{} & 
\multicolumn{1}{|p{3.19cm}}{} & 
\multicolumn{1}{|p{1.5cm}}{} & 
\multicolumn{1}{|p{2.91cm}}{} & 
\multicolumn{1}{|p{1.52cm}|}{} \\ 
\hline
\multicolumn{1}{|p{1.66cm}}{\textbf{O\textsubscript{3}}} & 
\multicolumn{1}{|p{2.85cm}}{} & 
\multicolumn{1}{|p{1.25cm}}{} & 
\multicolumn{1}{|p{2.86cm}}{} & 
\multicolumn{1}{|p{1.2cm}}{} & 
\multicolumn{1}{|p{3.19cm}}{} & 
\multicolumn{1}{|p{1.5cm}}{} & 
\multicolumn{1}{|p{2.91cm}}{} & 
\multicolumn{1}{|p{1.52cm}|}{} \\ 
\hline
\multicolumn{1}{|p{1.66cm}}{} & 
\multicolumn{1}{|p{2.85cm}}{cough(-0.34)} & 
\multicolumn{1}{|p{1.25cm}}{<.001} & 
\multicolumn{1}{|p{2.86cm}}{cough(-0.38)} & 
\multicolumn{1}{|p{1.2cm}}{<.001} & 
\multicolumn{1}{|p{3.19cm}}{cough(-0.41)} & 
\multicolumn{1}{|p{1.5cm}}{<.001} & 
\multicolumn{1}{|p{2.91cm}}{cough(-0.41)} & 
\multicolumn{1}{|p{1.52cm}|}{<.001} \\ 
\hline
\multicolumn{1}{|p{1.66cm}}{} & 
\multicolumn{1}{|p{2.85cm}}{bronchitis(-0.31)} & 
\multicolumn{1}{|p{1.25cm}}{<.001} & 
\multicolumn{1}{|p{2.86cm}}{bronchitis(-0.32)} & 
\multicolumn{1}{|p{1.2cm}}{<.001} & 
\multicolumn{1}{|p{3.19cm}}{bronchitis(-0.33)} & 
\multicolumn{1}{|p{1.5cm}}{<.001} & 
\multicolumn{1}{|p{2.91cm}}{bronchitis(-0.35)} & 
\multicolumn{1}{|p{1.52cm}|}{<.001} \\ 
\hline
\multicolumn{1}{|p{1.66cm}}{} & 
\multicolumn{1}{|p{2.85cm}}{traffic(0.26)} & 
\multicolumn{1}{|p{1.25cm}}{<.001} & 
\multicolumn{1}{|p{2.86cm}}{traffic(0.27)} & 
\multicolumn{1}{|p{1.2cm}}{<.001} & 
\multicolumn{1}{|p{3.19cm}}{traffic(0.26)} & 
\multicolumn{1}{|p{1.5cm}}{<.001} & 
\multicolumn{1}{|p{2.91cm}}{smoke(0.24)} & 
\multicolumn{1}{|p{1.52cm}|}{<.001} \\ 
\hline
\multicolumn{1}{|p{1.66cm}}{} & 
\multicolumn{1}{|p{2.85cm}}{smoke(0.23)} & 
\multicolumn{1}{|p{1.25cm}}{<.001} & 
\multicolumn{1}{|p{2.86cm}}{chest pain(-0.23)} & 
\multicolumn{1}{|p{1.2cm}}{<.001} & 
\multicolumn{1}{|p{3.19cm}}{chest pain(-0.23)} & 
\multicolumn{1}{|p{1.5cm}}{<.001} & 
\multicolumn{1}{|p{2.91cm}}{traffic(0.23)} & 
\multicolumn{1}{|p{1.52cm}|}{<.001} \\ 
\hline
\multicolumn{1}{|p{1.66cm}}{} & 
\multicolumn{1}{|p{2.85cm}}{snoring(0.22)} & 
\multicolumn{1}{|p{1.25cm}}{<.001} & 
\multicolumn{1}{|p{2.86cm}}{snoring(0.22)} & 
\multicolumn{1}{|p{1.2cm}}{<.001} & 
\multicolumn{1}{|p{3.19cm}}{smoke(0.22)} & 
\multicolumn{1}{|p{1.5cm}}{<.001} & 
\multicolumn{1}{|p{2.91cm}}{chest pain(-0.22)} & 
\multicolumn{1}{|p{1.52cm}|}{<.001} \\ 
\hline
\multicolumn{1}{|p{1.66cm}}{\textbf{NO\textsubscript{2}}} & 
\multicolumn{1}{|p{2.85cm}}{} & 
\multicolumn{1}{|p{1.25cm}}{} & 
\multicolumn{1}{|p{2.86cm}}{} & 
\multicolumn{1}{|p{1.2cm}}{} & 
\multicolumn{1}{|p{3.19cm}}{} & 
\multicolumn{1}{|p{1.5cm}}{} & 
\multicolumn{1}{|p{2.91cm}}{} & 
\multicolumn{1}{|p{1.52cm}|}{} \\ 
\hline
\multicolumn{1}{|p{1.66cm}}{} & 
\multicolumn{1}{|p{2.85cm}}{asthma(0.20)} & 
\multicolumn{1}{|p{1.25cm}}{<.001} & 
\multicolumn{1}{|p{2.86cm}}{sulfate(0.20)} & 
\multicolumn{1}{|p{1.2cm}}{<.001} & 
\multicolumn{1}{|p{3.19cm}}{sulfate(0.16)} & 
\multicolumn{1}{|p{1.5cm}}{.002} & 
\multicolumn{1}{|p{2.91cm}}{cough(0.16)} & 
\multicolumn{1}{|p{1.52cm}|}{.002} \\ 
\hline
\multicolumn{1}{|p{1.66cm}}{} & 
\multicolumn{1}{|p{2.85cm}}{sulfate(0.19)} & 
\multicolumn{1}{|p{1.25cm}}{<.001} & 
\multicolumn{1}{|p{2.86cm}}{bronchitis(0.16)} & 
\multicolumn{1}{|p{1.2cm}}{.002} & 
\multicolumn{1}{|p{3.19cm}}{bronchitis(0.15)} & 
\multicolumn{1}{|p{1.5cm}}{.005} & 
\multicolumn{1}{|p{2.91cm}}{copd(-0.16)} & 
\multicolumn{1}{|p{1.52cm}|}{.003} \\ 
\hline
\multicolumn{1}{|p{1.66cm}}{} & 
\multicolumn{1}{|p{2.85cm}}{cough(0.17)} & 
\multicolumn{1}{|p{1.25cm}}{<.001} & 
\multicolumn{1}{|p{2.86cm}}{inhaler(0.15)} & 
\multicolumn{1}{|p{1.2cm}}{.005} & 
\multicolumn{1}{|p{3.19cm}}{cough(0.14)} & 
\multicolumn{1}{|p{1.5cm}}{.008} & 
\multicolumn{1}{|p{2.91cm}}{bronchitis(0.14)} & 
\multicolumn{1}{|p{1.52cm}|}{.008} \\ 
\hline
\multicolumn{1}{|p{1.66cm}}{} & 
\multicolumn{1}{|p{2.85cm}}{bronchitis(0.17)} & 
\multicolumn{1}{|p{1.25cm}}{.001} & 
\multicolumn{1}{|p{2.86cm}}{cough(0.14)} & 
\multicolumn{1}{|p{1.2cm}}{.006} & 
\multicolumn{1}{|p{3.19cm}}{inhaler(0.11)} & 
\multicolumn{1}{|p{1.5cm}}{.03} & 
\multicolumn{1}{|p{2.91cm}}{wheezing(-0.12)} & 
\multicolumn{1}{|p{1.52cm}|}{.02} \\ 
\hline
\multicolumn{1}{|p{1.66cm}}{} & 
\multicolumn{1}{|p{2.85cm}}{inhaler(0.16)} & 
\multicolumn{1}{|p{1.25cm}}{.002} & 
\multicolumn{1}{|p{2.86cm}}{difficulty breathing (-0.12)} & 
\multicolumn{1}{|p{1.2cm}}{.02} & 
\multicolumn{1}{|p{3.19cm}}{headache(-0.11)} & 
\multicolumn{1}{|p{1.5cm}}{.03} & 
\multicolumn{1}{|p{2.91cm}}{headache(-0.10)} & 
\multicolumn{1}{|p{1.52cm}|}{.04} \\ 
\hline
\multicolumn{1}{|p{1.66cm}}{\textbf{PM\textsubscript{2.5}}} & 
\multicolumn{1}{|p{2.85cm}}{} & 
\multicolumn{1}{|p{1.25cm}}{} & 
\multicolumn{1}{|p{2.86cm}}{} & 
\multicolumn{1}{|p{1.2cm}}{} & 
\multicolumn{1}{|p{3.19cm}}{} & 
\multicolumn{1}{|p{1.5cm}}{} & 
\multicolumn{1}{|p{2.91cm}}{} & 
\multicolumn{1}{|p{1.52cm}|}{} \\ 
\hline
\multicolumn{1}{|p{1.66cm}}{} & 
\multicolumn{1}{|p{2.85cm}}{wildfires(0.14)} & 
\multicolumn{1}{|p{1.25cm}}{.009} & 
\multicolumn{1}{|p{2.86cm}}{copd(-0.15)} & 
\multicolumn{1}{|p{1.2cm}}{.005} & 
\multicolumn{1}{|p{3.19cm}}{air pollution(0.19)} & 
\multicolumn{1}{|p{1.5cm}}{<.001} & 
\multicolumn{1}{|p{2.91cm}}{air pollution(0.18)} & 
\multicolumn{1}{|p{1.52cm}|}{<.001} \\ 
\hline
\multicolumn{1}{|p{1.66cm}}{} & 
\multicolumn{1}{|p{2.85cm}}{copd(-0.11)} & 
\multicolumn{1}{|p{1.25cm}}{.03} & 
\multicolumn{1}{|p{2.86cm}}{wildfires(0.14)} & 
\multicolumn{1}{|p{1.2cm}}{.007} & 
\multicolumn{1}{|p{3.19cm}}{copd(-0.17)} & 
\multicolumn{1}{|p{1.5cm}}{.001} & 
\multicolumn{1}{|p{2.91cm}}{copd(-0.18)} & 
\multicolumn{1}{|p{1.52cm}|}{<.001} \\ 
\hline
\multicolumn{1}{|p{1.66cm}}{} & 
\multicolumn{1}{|p{2.85cm}}{snoring(0.11)} & 
\multicolumn{1}{|p{1.25cm}}{.03} & 
\multicolumn{1}{|p{2.86cm}}{air pollution(0.14)} & 
\multicolumn{1}{|p{1.2cm}}{.008} & 
\multicolumn{1}{|p{3.19cm}}{wildfires(0.14)} & 
\multicolumn{1}{|p{1.5cm}}{.009} & 
\multicolumn{1}{|p{2.91cm}}{wildfires(0.15)} & 
\multicolumn{1}{|p{1.52cm}|}{.004} \\ 
\hline
\multicolumn{1}{|p{1.66cm}}{} & 
\multicolumn{1}{|p{2.85cm}}{inhaler(0.10)} & 
\multicolumn{1}{|p{1.25cm}}{.06} & 
\multicolumn{1}{|p{2.86cm}}{asthma attack(0.11)} & 
\multicolumn{1}{|p{1.2cm}}{.04} & 
\multicolumn{1}{|p{3.19cm}}{respiratory illness(0.10)} & 
\multicolumn{1}{|p{1.5cm}}{.05} & 
\multicolumn{1}{|p{2.91cm}}{sulfate(-0.11)} & 
\multicolumn{1}{|p{1.52cm}|}{.03} \\ 
\hline
\multicolumn{1}{|p{1.66cm}}{} & 
\multicolumn{1}{|p{2.85cm}}{difficulty breathing (-0.09)} & 
\multicolumn{1}{|p{1.25cm}}{.08} & 
\multicolumn{1}{|p{2.86cm}}{respiratory illness (0.10)} & 
\multicolumn{1}{|p{1.2cm}}{.05} & 
\multicolumn{1}{|p{3.19cm}}{traffic(0.10)} & 
\multicolumn{1}{|p{1.5cm}}{.06} & 
\multicolumn{1}{|p{2.91cm}}{traffic(0.11)} & 
\multicolumn{1}{|p{1.52cm}|}{.04} \\ 
\hline
\multicolumn{1}{}{} \\ 
\end{tabular}
\end{adjustbox}
{\raggedright \textit{ $P^a$ }value, with n $=$ 366 \par}
\caption{Cross correlation of top five search terms with different lags for three pollutants in the Atlanta metropolitan area in 2016.}
\end{table}




\begin{table}[h]
\begin{adjustbox}{max width=\textwidth}
\begin{tabular}{p{2.79cm}p{3.61cm}p{2.94cm}p{2.94cm}p{2.95cm}p{2.79cm}p{3.61cm}p{2.94cm}p{2.94cm}p{2.95cm}}
\hline
\multicolumn{1}{p{2.79cm}}{Features} & 
\multicolumn{1}{p{3.61cm}}{Model } & 
\multicolumn{1}{p{2.94cm}}{O\textsubscript{3}} & 
\multicolumn{1}{p{2.94cm}}{NO\textsubscript{2}} & 
\multicolumn{1}{p{2.95cm}}{PM\textsubscript{2.5}} \\ 
\hhline{~~~~~}
\multicolumn{1}{p{2.79cm}}{} & 
\multicolumn{1}{p{3.61cm}}{} & 
\multicolumn{1}{p{2.94cm}}{Accuracy$\%$ \newline
(F1$\%$)} & 
\multicolumn{1}{p{2.94cm}}{Accuracy$\%$ \newline
(F1$\%$)} & 
\multicolumn{1}{p{2.95cm}}{Accuracy$\%$ \newline
(F1$\%$)} \\ 
\hline
\multicolumn{1}{p{2.79cm}}{\textbf{No Prior Knowledge}} & 
\multicolumn{1}{p{3.61cm}}{} & 
\multicolumn{1}{p{2.94cm}}{} & 
\multicolumn{1}{p{2.94cm}}{} & 
\multicolumn{1}{p{2.95cm}}{} \\ 
\multicolumn{1}{p{2.79cm}}{} & 
\multicolumn{1}{p{3.61cm}}{All Positives } & 
\multicolumn{1}{p{2.94cm}}{\centering
13.46 (23.73)} & 
\multicolumn{1}{p{2.94cm}}{\centering
13.18 (23.28)} & 
\multicolumn{1}{p{2.95cm}}{\centering
7.35 (13.69)} \\ 
\multicolumn{1}{p{2.79cm}}{} & 
\multicolumn{1}{p{3.61cm}}{All Negatives} & 
\multicolumn{1}{p{2.94cm}}{\centering
86.54 (0.0)} & 
\multicolumn{1}{p{2.94cm}}{\centering
86.82 (0.0)} & 
\multicolumn{1}{p{2.95cm}}{\centering
92.65 (0.0)} \\ 
\multicolumn{1}{p{2.79cm}}{} & 
\multicolumn{1}{p{3.61cm}}{Random (Prob$=$0.5)} & 
\multicolumn{1}{p{2.94cm}}{\centering
50.29 (20.63)} & 
\multicolumn{1}{p{2.94cm}}{\centering
50.56 (20.68)} & 
\multicolumn{1}{p{2.95cm}}{\centering
50.65 (12.67)} \\ 
\multicolumn{1}{p{2.79cm}}{\textbf{Search}} & 
\multicolumn{1}{p{3.61cm}}{} & 
\multicolumn{1}{p{2.94cm}}{} & 
\multicolumn{1}{p{2.94cm}}{} & 
\multicolumn{1}{p{2.95cm}}{} \\ 
\multicolumn{1}{p{2.79cm}}{} & 
\multicolumn{1}{p{3.61cm}}{LR } & 
\multicolumn{1}{p{2.94cm}}{\centering
36.93 (17.77)} & 
\multicolumn{1}{p{2.94cm}}{\centering
53.97 (24.17)} & 
\multicolumn{1}{p{2.95cm}}{\centering
78.29 (10.72)} \\ 
\hhline{~~~~~}
\multicolumn{1}{p{2.79cm}}{} & 
\multicolumn{1}{p{3.61cm}}{RF } & 
\multicolumn{1}{p{2.94cm}}{\centering
33.53 (23.36)} & 
\multicolumn{1}{p{2.94cm}}{\centering
55.22 (18.1)} & 
\multicolumn{1}{p{2.95cm}}{\centering
\textit{92.65\footnote{   This high accuracy is simply due to class imbalance; this model always predicts negative class and the corresponding F1 score is zero.}(0.0)}} \\ 
\hhline{~~~~~}
\multicolumn{1}{p{2.79cm}}{} & 
\multicolumn{1}{p{3.61cm}}{LSTM } & 
\multicolumn{1}{p{2.94cm}}{\centering
46.73 (23.63)} & 
\multicolumn{1}{p{2.94cm}}{\centering
69.68 (21.62)} & 
\multicolumn{1}{p{2.95cm}}{\centering
89.96 (7.58)} \\ 
\hhline{~~~~~}
\multicolumn{1}{p{2.79cm}}{} & 
\multicolumn{1}{p{3.61cm}}{LSTM-GloVe } & 
\multicolumn{1}{p{2.94cm}}{\centering
53.23 (28.45)} & 
\multicolumn{1}{p{2.94cm}}{\centering
63.44 (27.4)} & 
\multicolumn{1}{p{2.95cm}}{\centering
90.09 (3.73)} \\ 
\hhline{~~~~~}
\multicolumn{1}{p{2.79cm}}{} & 
\multicolumn{1}{p{3.61cm}}{LSTM-GloVe w/ STE  } & 
\multicolumn{1}{p{2.94cm}}{\centering
69.17 (28.04)} & 
\multicolumn{1}{p{2.94cm}}{\centering
46.85 (26.51)} & 
\multicolumn{1}{p{2.95cm}}{\centering
91.73 (1.31)} \\ 
\hhline{~~~~~}
\multicolumn{1}{p{2.79cm}}{} & 
\multicolumn{1}{p{3.61cm}}{DL-LSTM   } & 
\multicolumn{1}{p{2.94cm}}{\centering
62.46 (30.4)} & 
\multicolumn{1}{p{2.94cm}}{\centering
65.99 (26.19)} & 
\multicolumn{1}{p{2.95cm}}{\centering
88.61 (7.97)} \\ 
\multicolumn{1}{p{2.79cm}}{} & 
\multicolumn{1}{p{3.61cm}}{DL-LSTM w/ STE } & 
\multicolumn{1}{p{2.94cm}}{\centering
69.61 (32.44)} & 
\multicolumn{1}{p{2.94cm}}{\centering
56.84 (27.7)} & 
\multicolumn{1}{p{2.95cm}}{\centering
87.59 (6.99)} \\ 
\multicolumn{1}{p{2.79cm}}{\textbf{Met}} & 
\multicolumn{1}{p{3.61cm}}{} & 
\multicolumn{1}{p{2.94cm}}{} & 
\multicolumn{1}{p{2.94cm}}{} & 
\multicolumn{1}{p{2.95cm}}{} \\ 
\hhline{~~~~~}
\multicolumn{1}{p{2.79cm}}{} & 
\multicolumn{1}{p{3.61cm}}{LR } & 
\multicolumn{1}{p{2.94cm}}{\centering
62.57 (39.81)} & 
\multicolumn{1}{p{2.94cm}}{\centering
63.64 (37.25)} & 
\multicolumn{1}{p{2.95cm}}{\centering
58.58 (22)} \\ 
\hhline{~~~~~}
\multicolumn{1}{p{2.79cm}}{} & 
\multicolumn{1}{p{3.61cm}}{RF } & 
\multicolumn{1}{p{2.94cm}}{\centering
78.76 (50.59)} & 
\multicolumn{1}{p{2.94cm}}{\centering
71.77 (39.88)} & 
\multicolumn{1}{p{2.95cm}}{\centering
73.78 (24.67)} \\ 
\hhline{~~~~~}
\multicolumn{1}{p{2.79cm}}{} & 
\multicolumn{1}{p{3.61cm}}{LSTM } & 
\multicolumn{1}{p{2.94cm}}{\centering
76.54 (48.29)} & 
\multicolumn{1}{p{2.94cm}}{\centering
72.52 (41.27)} & 
\multicolumn{1}{p{2.95cm}}{\centering
67.89 (24.69)} \\ 
\hhline{~~~~~}
\multicolumn{1}{p{2.79cm}}{\textbf{Met+Search}} & 
\multicolumn{1}{p{3.61cm}}{} & 
\multicolumn{1}{p{2.94cm}}{} & 
\multicolumn{1}{p{2.94cm}}{} & 
\multicolumn{1}{p{2.95cm}}{} \\ 
\hhline{~~~~~}
\multicolumn{1}{p{2.79cm}}{} & 
\multicolumn{1}{p{3.61cm}}{LR } & 
\multicolumn{1}{p{2.94cm}}{\centering
55.99 (36.56)} & 
\multicolumn{1}{p{2.94cm}}{\centering
62 (36.25)} & 
\multicolumn{1}{p{2.95cm}}{\centering
61.25 (21.5)} \\ 
\hhline{~~~~~}
\multicolumn{1}{p{2.79cm}}{} & 
\multicolumn{1}{p{3.61cm}}{RF } & 
\multicolumn{1}{p{2.94cm}}{\centering
81.39 (45.35)} & 
\multicolumn{1}{p{2.94cm}}{\centering
73.77 (38.71)} & 
\multicolumn{1}{p{2.95cm}}{\centering
87.96 (23.78)} \\ 
\hhline{~~~~~}
\multicolumn{1}{p{2.79cm}}{} & 
\multicolumn{1}{p{3.61cm}}{LSTM } & 
\multicolumn{1}{p{2.94cm}}{\centering
78.18 (47.65)} & 
\multicolumn{1}{p{2.94cm}}{\centering
77.75 (40.31)} & 
\multicolumn{1}{p{2.95cm}}{\centering
88.14 (21.29)} \\ 
\hhline{~~~~~}
\multicolumn{1}{p{2.79cm}}{} & 
\multicolumn{1}{p{3.61cm}}{LSTM-GloVe  } & 
\multicolumn{1}{p{2.94cm}}{\centering
80.04 (49.37)} & 
\multicolumn{1}{p{2.94cm}}{\centering
72.75 (40.35)} & 
\multicolumn{1}{p{2.95cm}}{\centering
85.38 (26.99)} \\ 
\hhline{~~~~~}
\multicolumn{1}{p{2.79cm}}{} & 
\multicolumn{1}{p{3.61cm}}{LSTM-GloVe w/ STE  } & 
\multicolumn{1}{p{2.94cm}}{\centering
81.85 (50.71)} & 
\multicolumn{1}{p{2.94cm}}{\centering
74.21 (41.49)} & 
\multicolumn{1}{p{2.95cm}}{\centering
85.42 (26.13)} \\ 
\hhline{~~~~~}
\multicolumn{1}{p{2.79cm}}{} & 
\multicolumn{1}{p{3.61cm}}{DL-LSTM } & 
\multicolumn{1}{p{2.94cm}}{\centering
77.97 (48.94)} & 
\multicolumn{1}{p{2.94cm}}{\centering
74.81 (40.53)} & 
\multicolumn{1}{p{2.95cm}}{\centering
84.94 (24.07)} \\ 
\hhline{~~~~~}
\multicolumn{1}{p{2.79cm}}{} & 
\multicolumn{1}{p{3.61cm}}{DL-LSTM w/ STE } & 
\multicolumn{1}{p{2.94cm}}{\centering
80.16 (49.32)} & 
\multicolumn{1}{p{2.94cm}}{72.99 (40.34)} & 
\multicolumn{1}{p{2.95cm}}{\centering
87.04 (21.32)} \\ 
\hhline{~~~~~}
\multicolumn{1}{p{2.79cm}}{\centering
\textbf{Met +Pol}} & 
\multicolumn{1}{p{3.61cm}}{} & 
\multicolumn{1}{p{2.94cm}}{} & 
\multicolumn{1}{p{2.94cm}}{} & 
\multicolumn{1}{p{2.95cm}}{} \\ 
\hhline{~~~~~}
\multicolumn{1}{p{2.79cm}}{} & 
\multicolumn{1}{p{3.61cm}}{LR } & 
\multicolumn{1}{p{2.94cm}}{\centering
67.38 (44.61)} & 
\multicolumn{1}{p{2.94cm}}{\centering
70.05 (44.09)} & 
\multicolumn{1}{p{2.95cm}}{\centering
74.45 (32.82)} \\ 
\hhline{~~~~~}
\multicolumn{1}{p{2.79cm}}{} & 
\multicolumn{1}{p{3.61cm}}{RF } & 
\multicolumn{1}{p{2.94cm}}{\centering
82.81 (57.23)} & 
\multicolumn{1}{p{2.94cm}}{\centering
80.35 (51.24)} & 
\multicolumn{1}{p{2.95cm}}{\centering
86.45 (40.63)} \\ 
\hhline{~~~~~}
\multicolumn{1}{p{2.79cm}}{} & 
\multicolumn{1}{p{3.61cm}}{LSTM } & 
\multicolumn{1}{p{2.94cm}}{86.97 (63.01)} & 
\multicolumn{1}{p{2.94cm}}{\centering
84.64 (55.59)} & 
\multicolumn{1}{p{2.95cm}}{\centering
85.25 (43.19)} \\ 
\hhline{~~~~~}
\multicolumn{1}{p{2.79cm}}{\centering
\textbf{Met+Pol+Search}} & 
\multicolumn{1}{p{3.61cm}}{} & 
\multicolumn{1}{p{2.94cm}}{} & 
\multicolumn{1}{p{2.94cm}}{} & 
\multicolumn{1}{p{2.95cm}}{} \\ 
\hhline{~~~~~}
\multicolumn{1}{p{2.79cm}}{} & 
\multicolumn{1}{p{3.61cm}}{LR } & 
\multicolumn{1}{p{2.94cm}}{\centering
66.91 (43.71)} & 
\multicolumn{1}{p{2.94cm}}{\centering
69.13 (43.6)} & 
\multicolumn{1}{p{2.95cm}}{\centering
74.45 (32.82)} \\ 
\hhline{~~~~~}
\multicolumn{1}{p{2.79cm}}{} & 
\multicolumn{1}{p{3.61cm}}{RF } & 
\multicolumn{1}{p{2.94cm}}{\centering
82.76 (55.91)} & 
\multicolumn{1}{p{2.94cm}}{\centering
78.91 (47.72)} & 
\multicolumn{1}{p{2.95cm}}{\centering
89.43 (37.57)} \\ 
\hhline{~~~~~}
\multicolumn{1}{p{2.79cm}}{} & 
\multicolumn{1}{p{3.61cm}}{LSTM } & 
\multicolumn{1}{p{2.94cm}}{87.11 (61.54)} & 
\multicolumn{1}{p{2.94cm}}{\centering
84.71 (54.02)} & 
\multicolumn{1}{p{2.95cm}}{\centering
90.74 (44.81)} \\ 
\hhline{~~~~~}
\multicolumn{1}{p{2.79cm}}{} & 
\multicolumn{1}{p{3.61cm}}{LSTM-GloVe  } & 
\multicolumn{1}{p{2.94cm}}{\centering
87.94 (63.81)} & 
\multicolumn{1}{p{2.94cm}}{\centering
82.98 (53.78)} & 
\multicolumn{1}{p{2.95cm}}{\centering
88.19 (46.55)} \\ 
\hhline{~~~~~}
\multicolumn{1}{p{2.79cm}}{} & 
\multicolumn{1}{p{3.61cm}}{LSTM-GloVe w/ STE  } & 
\multicolumn{1}{p{2.94cm}}{\centering
87.63 (63.83)} & 
\multicolumn{1}{p{2.94cm}}{\centering
83.81 (54.59)} & 
\multicolumn{1}{p{2.95cm}}{\centering
88.24 (46.51)} \\ 
\hhline{~~~~~}
\multicolumn{1}{p{2.79cm}}{} & 
\multicolumn{1}{p{3.61cm}}{DL-LSTM   } & 
\multicolumn{1}{p{2.94cm}}{\centering
87.30 (63.02)} & 
\multicolumn{1}{p{2.94cm}}{\centering
82.65 (53.65)} & 
\multicolumn{1}{p{2.95cm}}{\centering
89.66 (47.35)} \\ 
\hhline{~~~~~}
\multicolumn{1}{p{2.79cm}}{ } & 
\multicolumn{1}{p{3.61cm}}{DL-LSTM w/ STE } & 
\multicolumn{1}{p{2.94cm}}{\centering
87.60 (63.61)} & 
\multicolumn{1}{p{2.94cm}}{\centering
83.40 (53.58)} & 
\multicolumn{1}{p{2.95cm}}{\centering
89.25 (46.59)} \\ 
\hline
\end{tabular}
\end{adjustbox}
\caption{Accuracy and F1 score of the LR, RF, and LSTM models for detecting elevated pollution across 10 major U.S. cities, for varying input feature combinations: no prior knowledge, search data only (Search), meteorological data only (Met); meteorological data and search data (Met +Search), meteorological data and historical pollutant concentration (Met +Pol) and all input features (Met +Pol+Search).}
\end{table}

\vspace{1\baselineskip}
\paragraph{Using Search Data and Meteorological Data:}

When meteorological data is available, we investigated the feasibility of using meteorological data with/without search activity data to nowcast air pollution under this condition. As shown in the ``Met" and ``Met +Search" sections of (Table 5), the inclusion of Web search data improves the nowcasting accuracy for all three pollutants. In addition, the LSTM-GloVe w/ STE model achieves the highest F1 score (50.71$\%$ for O\textsubscript{3}, 41.49$\%$ for NO\textsubscript{2}) for detecting O\textsubscript{3 }and NO\textsubscript{2} pollution. The LSTM-GloVe w/o STE model achieves the highest F1 score (26.99$\%$) for detecting PM\textsubscript{2.5} pollution.

\paragraph{Using Search Data, Meteorological Data and Historical Pollutant Concentration:}

When historical pollution concentration is available, search activity data is added as auxiliary data to both meteorological data and historical pollution data. As shown in the ``Met+Pol" and ``Met+Pol+Search" sections of (Table 5), the inclusion of Web search data improves the nowcasting accuracy for O\textsubscript{3} and PM\textsubscript{2.5}. However, for NO\textsubscript{2, }the inclusion of Web search data does not improve the nowcasting accuracy, which indicates increases in NO\textsubscript{2} concentrations may not be directly noticeable by people sufficiently to increase their search interest. This difference in performance for different pollutants and locales merits further investigation.

\subsubsection*{City-level Analysis of Ozone Pollution Prediction}

We investigated the potential of using search interest and meteorological data to replace ground-based ozone sensor data for ozone pollution prediction in individual cities. As shown in (Table 6), including search interest data (Met+Search) to augment purely meteorological data (Met) increases both accuracy and F1 metrics for most cities. While these metrics are not reaching the performance when the ground-level pollution sensors are available (Met+Pol), at least for two of the major MSAs (Philadelphia and Houston), 


\begin{figure}[h]
\centering
\begin{subfigure}[b]{0.45\textwidth}
\centering
\includegraphics[width=\textwidth]{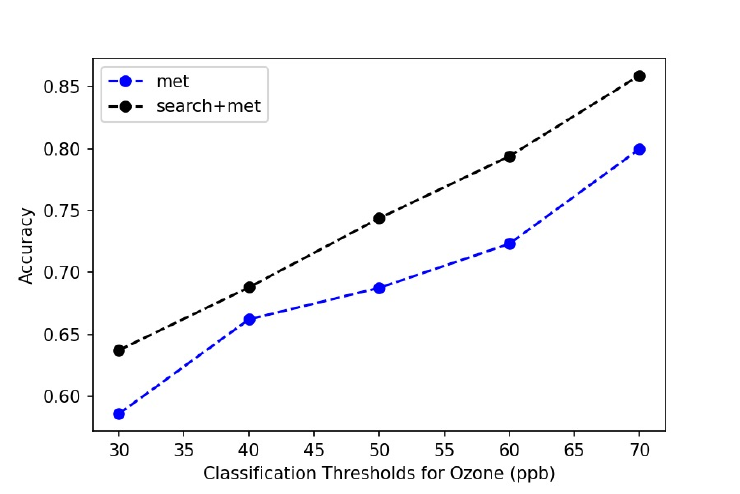}
\end{subfigure}
\hfill
\begin{subfigure}[b]{0.45\textwidth}
\centering
\includegraphics[width=\textwidth]{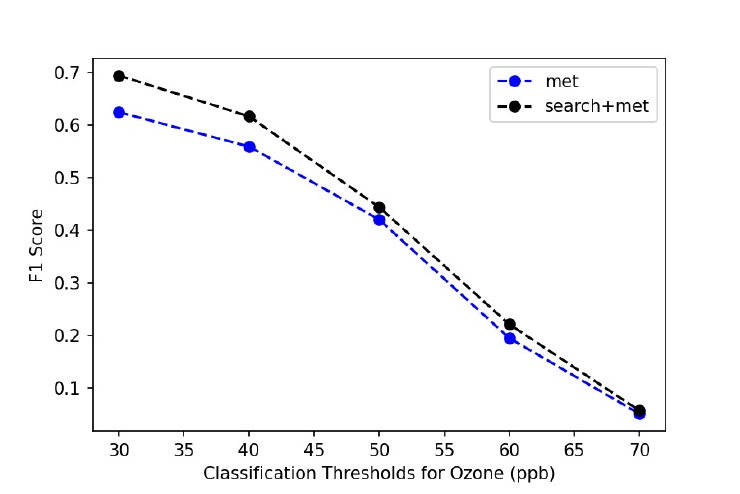}
\end{subfigure}
\caption{Accuracy (left figure) and F1 score (right figure) for detecting Ozone pollution on various classification thresholds, with Met (LSTM model) and Met+Search (DL-LSTM w/ STE) as features.}
\end{figure}

\begin{figure}[h]
\centering
\begin{subfigure}[b]{0.45\textwidth}
\centering
\includegraphics[width=\textwidth]{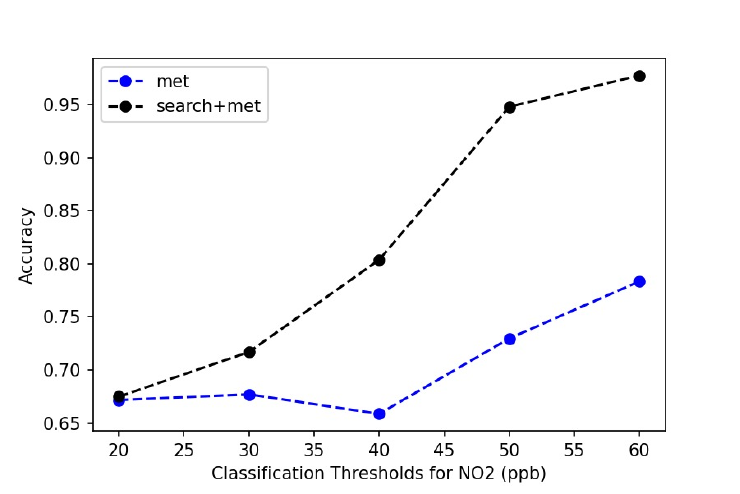}
\end{subfigure}
\hfill
\begin{subfigure}[b]{0.45\textwidth}
\centering
\includegraphics[width=\textwidth]{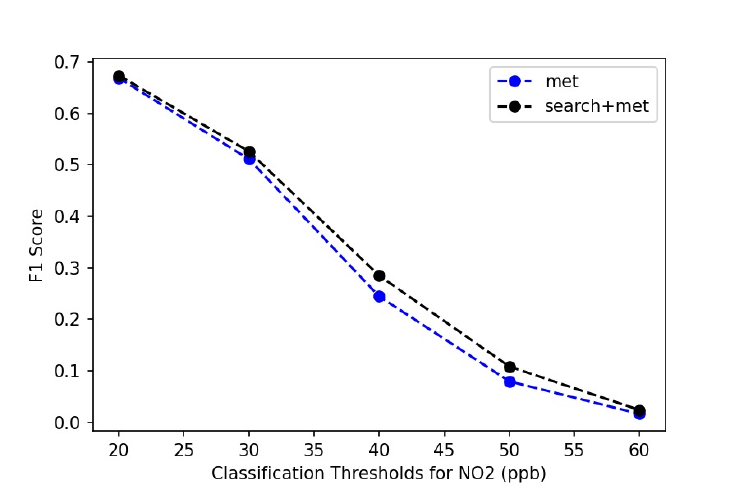}
\end{subfigure}
\caption{Accuracy (left figure) and F1 score (right figure) for detecting NO\textsubscript{2} pollution on various classification thresholds, with Met (LSTM model) and Met+Search (DL-LSTM w/ STE) as features.}
\end{figure}
\begin{figure}[h]
\centering
\begin{subfigure}[b]{0.45\textwidth}
\centering
\includegraphics[width=\textwidth]{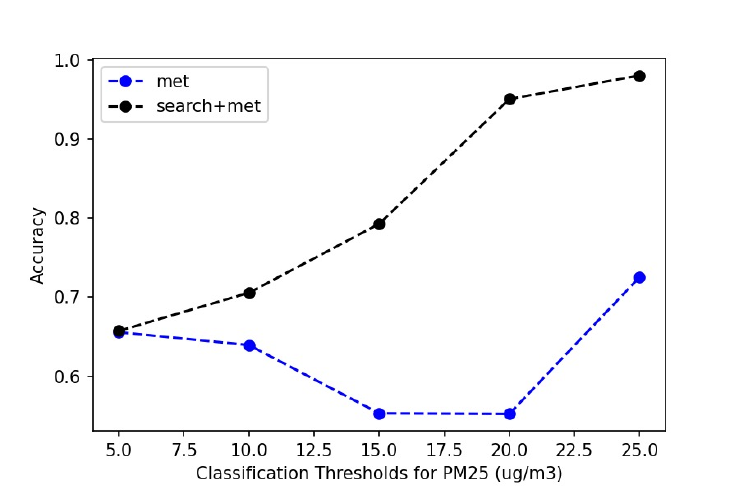}
\end{subfigure}
\hfill
\begin{subfigure}[b]{0.45\textwidth}
\centering
\includegraphics[width=\textwidth]{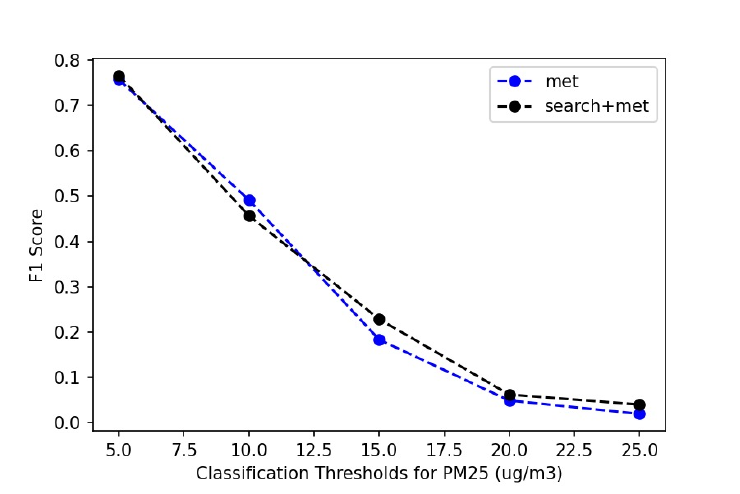}
\end{subfigure}
\caption{Accuracy (left figure) and F1 score (right figure) for detecting PM\textsubscript{2.5} pollution on various classification thresholds, with Met (LSTM model) and Met+Search (DL-LSTM w/ STE) as features.}
\end{figure}

\begin{table}[h]
\begin{adjustbox}{max width=\textwidth}
\begin{tabular}{p{0.42cm}p{1.72cm}p{1.31cm}p{1.31cm}p{1.31cm}p{1.31cm}p{1.31cm}p{1.31cm}p{1.31cm}p{1.31cm}p{1.31cm}p{1.31cm}p{0.42cm}p{1.72cm}p{1.31cm}p{1.31cm}p{1.31cm}p{1.31cm}p{1.31cm}p{1.31cm}p{1.31cm}p{1.31cm}p{1.31cm}p{1.31cm}}
\hline
\multicolumn{2}{|p{2.14cm}}{Features} & 
\multicolumn{1}{|p{1.72cm}}{ L.A. } & 
\multicolumn{1}{|p{1.31cm}}{ DC } & 
\multicolumn{1}{|p{1.31cm}}{ PHILA} & 
\multicolumn{1}{|p{1.31cm}}{ DTX} & 
\multicolumn{1}{|p{1.31cm}}{ ATL } & 
\multicolumn{1}{|p{1.31cm}}{ BOS } & 
\multicolumn{1}{|p{1.31cm}}{ NY } & 
\multicolumn{1}{|p{1.31cm}}{ MIA } & 
\multicolumn{1}{|p{1.31cm}}{ CHI } & 
\multicolumn{1}{|p{1.31cm}|}{ HOU } \\ 
\hline
\multicolumn{2}{|p{2.14cm}}{} & 
\multicolumn{1}{|p{1.72cm}}{} & 
\multicolumn{1}{|p{1.31cm}}{} & 
\multicolumn{1}{|p{1.31cm}}{} & 
\multicolumn{1}{|p{1.31cm}}{} & 
\multicolumn{1}{|p{1.31cm}}{} & 
\multicolumn{1}{|p{1.31cm}}{} & 
\multicolumn{1}{|p{1.31cm}}{} & 
\multicolumn{1}{|p{1.31cm}}{} & 
\multicolumn{1}{|p{1.31cm}}{} & 
\multicolumn{1}{|p{1.31cm}|}{} \\ 
\hline
\multicolumn{2}{|p{2.14cm}}{  Accuracy $\%$ } & 
\multicolumn{1}{|p{1.72cm}}{} & 
\multicolumn{1}{|p{1.31cm}}{} & 
\multicolumn{1}{|p{1.31cm}}{} & 
\multicolumn{1}{|p{1.31cm}}{} & 
\multicolumn{1}{|p{1.31cm}}{} & 
\multicolumn{1}{|p{1.31cm}}{} & 
\multicolumn{1}{|p{1.31cm}}{} & 
\multicolumn{1}{|p{1.31cm}}{} & 
\multicolumn{1}{|p{1.31cm}}{} & 
\multicolumn{1}{|p{1.31cm}|}{} \\ 
\hline
\multicolumn{1}{|p{0.42cm}}{} & 
\multicolumn{1}{|p{1.72cm}}{Met} & 
\multicolumn{1}{|p{1.31cm}}{\centering
72.6} & 
\multicolumn{1}{|p{1.31cm}}{\centering
77.4} & 
\multicolumn{1}{|p{1.31cm}}{\centering
83.29} & 
\multicolumn{1}{|p{1.31cm}}{\centering
83.42} & 
\multicolumn{1}{|p{1.31cm}}{\centering
83.56} & 
\multicolumn{1}{|p{1.31cm}}{\centering
75.62} & 
\multicolumn{1}{|p{1.31cm}}{\centering
68.36} & 
\multicolumn{1}{|p{1.31cm}}{\centering
58.09} & 
\multicolumn{1}{|p{1.31cm}}{\centering
76.71} & 
\multicolumn{1}{|p{1.31cm}|}{\centering
85.89} \\ 
\hline
\multicolumn{1}{|p{0.42cm}}{} & 
\multicolumn{1}{|p{1.72cm}}{Met +Search } & 
\multicolumn{1}{|p{1.31cm}}{\centering
76.71} & 
\multicolumn{1}{|p{1.31cm}}{\centering
80.68} & 
\multicolumn{1}{|p{1.31cm}}{\centering
87.4} & 
\multicolumn{1}{|p{1.31cm}}{\centering
79.86} & 
\multicolumn{1}{|p{1.31cm}}{\centering
83.84} & 
\multicolumn{1}{|p{1.31cm}}{\centering
78.63} & 
\multicolumn{1}{|p{1.31cm}}{\centering
74.93} & 
\multicolumn{1}{|p{1.31cm}}{\centering
69.29} & 
\multicolumn{1}{|p{1.31cm}}{\centering
80} & 
\multicolumn{1}{|p{1.31cm}|}{\centering
90.14} \\ 
\hline
\multicolumn{1}{|p{0.42cm}}{} & 
\multicolumn{1}{|p{1.72cm}}{Met +Pol} & 
\multicolumn{1}{|p{1.31cm}}{\centering
85.89} & 
\multicolumn{1}{|p{1.31cm}}{\centering
86.99} & 
\multicolumn{1}{|p{1.31cm}}{\centering
89.04} & 
\multicolumn{1}{|p{1.31cm}}{\centering
89.04} & 
\multicolumn{1}{|p{1.31cm}}{\centering
88.22} & 
\multicolumn{1}{|p{1.31cm}}{\centering
84.66} & 
\multicolumn{1}{|p{1.31cm}}{\centering
86.85} & 
\multicolumn{1}{|p{1.31cm}}{\centering
82.02} & 
\multicolumn{1}{|p{1.31cm}}{\centering
86.85} & 
\multicolumn{1}{|p{1.31cm}|}{\centering
90} \\ 
\hline
\multicolumn{2}{|p{2.14cm}}{ F1 $\%$ } & 
\multicolumn{1}{|p{1.72cm}}{} & 
\multicolumn{1}{|p{1.31cm}}{} & 
\multicolumn{1}{|p{1.31cm}}{} & 
\multicolumn{1}{|p{1.31cm}}{} & 
\multicolumn{1}{|p{1.31cm}}{} & 
\multicolumn{1}{|p{1.31cm}}{} & 
\multicolumn{1}{|p{1.31cm}}{} & 
\multicolumn{1}{|p{1.31cm}}{} & 
\multicolumn{1}{|p{1.31cm}}{} & 
\multicolumn{1}{|p{1.31cm}|}{} \\ 
\hline
\multicolumn{1}{|p{0.42cm}}{} & 
\multicolumn{1}{|p{1.72cm}}{ Met} & 
\multicolumn{1}{|p{1.31cm}}{\centering
51.69} & 
\multicolumn{1}{|p{1.31cm}}{\centering
48.28} & 
\multicolumn{1}{|p{1.31cm}}{\centering
53.79} & 
\multicolumn{1}{|p{1.31cm}}{\centering
53.28} & 
\multicolumn{1}{|p{1.31cm}}{\centering
48.72} & 
\multicolumn{1}{|p{1.31cm}}{\centering
46.06} & 
\multicolumn{1}{|p{1.31cm}}{\centering
44.07} & 
\multicolumn{1}{|p{1.31cm}}{\centering
32.52} & 
\multicolumn{1}{|p{1.31cm}}{\centering
56.19} & 
\multicolumn{1}{|p{1.31cm}|}{\centering
57.26} \\ 
\hline
\multicolumn{1}{|p{0.42cm}}{} & 
\multicolumn{1}{|p{1.72cm}}{Met +Search  } & 
\multicolumn{1}{|p{1.31cm}}{\centering
54.3} & 
\multicolumn{1}{|p{1.31cm}}{\centering
50.53} & 
\multicolumn{1}{|p{1.31cm}}{\centering
58.56} & 
\multicolumn{1}{|p{1.31cm}}{\centering
41.9} & 
\multicolumn{1}{|p{1.31cm}}{\centering
42.72} & 
\multicolumn{1}{|p{1.31cm}}{\centering
48} & 
\multicolumn{1}{|p{1.31cm}}{\centering
47.86} & 
\multicolumn{1}{|p{1.31cm}}{\centering
35.84} & 
\multicolumn{1}{|p{1.31cm}}{\centering
57.56} & 
\multicolumn{1}{|p{1.31cm}|}{\centering
59.09} \\ 
\hline
\multicolumn{1}{|p{0.42cm}}{} & 
\multicolumn{1}{|p{1.72cm}}{Met +Pol } & 
\multicolumn{1}{|p{1.31cm}}{\centering
68.11} & 
\multicolumn{1}{|p{1.31cm}}{\centering
60.58} & 
\multicolumn{1}{|p{1.31cm}}{\centering
64.29} & 
\multicolumn{1}{|p{1.31cm}}{\centering
64.6} & 
\multicolumn{1}{|p{1.31cm}}{\centering
56.12} & 
\multicolumn{1}{|p{1.31cm}}{\centering
55.56} & 
\multicolumn{1}{|p{1.31cm}}{\centering
63.64} & 
\multicolumn{1}{|p{1.31cm}}{\centering
55.48} & 
\multicolumn{1}{|p{1.31cm}}{\centering
70.73} & 
\multicolumn{1}{|p{1.31cm}|}{\centering
67.26} \\ 
\hline
\end{tabular}
\end{adjustbox}
\caption{City-level accuracy and F1 Score for detecting elevated O\textsubscript{3} pollution in 10 U.S. cities, with Met (LSTM model), Met+Search (DL-LSTM w/ STE) and Met+Pol (LSTM model) as features.}
\end{table}

\begin{table}[h]
\begin{adjustbox}{max width=\textwidth}
\begin{tabular}{p{1.52cm}p{1.52cm}p{1.52cm}p{1.52cm}p{1.52cm}p{1.52cm}p{1.52cm}p{1.52cm}p{1.52cm}p{1.52cm}p{1.52cm}p{1.52cm}p{1.52cm}p{1.52cm}p{1.52cm}p{1.52cm}p{1.52cm}p{1.52cm}p{1.52cm}p{1.52cm}}
\hline
\multicolumn{10}{|p{15.199999999999998cm}|}{\centering
Search Term (Spearman's correlation, lag $=$ 1)} \\ 
\hline
\multicolumn{1}{|p{1.52cm}}{\centering
L.A.} & 
\multicolumn{1}{|p{1.52cm}}{\centering
DC} & 
\multicolumn{1}{|p{1.52cm}}{\centering
PHILA} & 
\multicolumn{1}{|p{1.52cm}}{\centering
DTX} & 
\multicolumn{1}{|p{1.52cm}}{\centering
ATL} & 
\multicolumn{1}{|p{1.52cm}}{\centering
BOS} & 
\multicolumn{1}{|p{1.52cm}}{\centering
NY} & 
\multicolumn{1}{|p{1.52cm}}{\centering
MIA} & 
\multicolumn{1}{|p{1.52cm}}{\centering
CHI} & 
\multicolumn{1}{|p{1.52cm}|}{\centering
HOU} \\ 
\hline
\multicolumn{1}{|p{1.52cm}}{} & 
\multicolumn{1}{|p{1.52cm}}{} & 
\multicolumn{1}{|p{1.52cm}}{} & 
\multicolumn{1}{|p{1.52cm}}{} & 
\multicolumn{1}{|p{1.52cm}}{} & 
\multicolumn{1}{|p{1.52cm}}{} & 
\multicolumn{1}{|p{1.52cm}}{} & 
\multicolumn{1}{|p{1.52cm}}{} & 
\multicolumn{1}{|p{1.52cm}}{} & 
\multicolumn{1}{|p{1.52cm}|}{} \\ 
\hline
\multicolumn{1}{|p{1.52cm}}{\centering
{\scriptsize cough \\ (-0.40)}} & 
\multicolumn{1}{|p{1.52cm}}{\centering
{\scriptsize bronchitis \\ (-0.25)}} & 
\multicolumn{1}{|p{1.52cm}}{\centering
{\scriptsize cough \\ (-0.33)}} & 
\multicolumn{1}{|p{1.52cm}}{\centering
{\scriptsize cough \\ (-0.25)}} & 
\multicolumn{1}{|p{1.52cm}}{\centering
{\scriptsize bronchitis \\ (-0.14)}} & 
\multicolumn{1}{|p{1.52cm}}{\centering
{\scriptsize smoke \\ (-0.11)}} & 
\multicolumn{1}{|p{1.52cm}}{\centering
{\scriptsize bronchitis \\ (-0.31)}} & 
\multicolumn{1}{|p{1.52cm}}{\centering
{\scriptsize bronchitis \\ (0.14)}} & 
\multicolumn{1}{|p{1.52cm}}{\centering
{\scriptsize wildfires \\ (0.18)}} & 
\multicolumn{1}{|p{1.52cm}|}{\centering
{\scriptsize ozone \\ (0.12)}} \\ 
\hline
\multicolumn{1}{|p{1.52cm}}{\centering
{\scriptsize bronchitis \\ (-0.33)}} & 
\multicolumn{1}{|p{1.52cm}}{\centering
{\scriptsize cough \\ (-0.25)}} & 
\multicolumn{1}{|p{1.52cm}}{\centering
{\scriptsize traffic \\ (0.27)}} & 
\multicolumn{1}{|p{1.52cm}}{\centering
{\scriptsize bronchitis \\ (-0.24)}} & 
\multicolumn{1}{|p{1.52cm}}{\centering
{\scriptsize cough \\ (-0.11)}} & 
\multicolumn{1}{|p{1.52cm}}{\centering
{\scriptsize haze \\ (-0.07)}} & 
\multicolumn{1}{|p{1.52cm}}{\centering
{\scriptsize traffic \\ (0.29)}} & 
\multicolumn{1}{|p{1.52cm}}{\centering
{\scriptsize air pollution \\ (0.13)}} & 
\multicolumn{1}{|p{1.52cm}}{\centering
{\scriptsize smoke \\ (0.08)}} & 
\multicolumn{1}{|p{1.52cm}|}{\centering
{\scriptsize air pollution \\ (0.12)}} \\ 
\hline
\multicolumn{1}{|p{1.52cm}}{\centering
{\scriptsize wildfires \\ (0.24)}} & 
\multicolumn{1}{|p{1.52cm}}{\centering
{\scriptsize coughing \\ (-0.19)}} & 
\multicolumn{1}{|p{1.52cm}}{\centering
{\scriptsize bronchitis \\ (-0.20)}} & 
\multicolumn{1}{|p{1.52cm}}{\centering
{\scriptsize ozone \\ (0.17)}} & 
\multicolumn{1}{|p{1.52cm}}{\centering
{\scriptsize chest pain \\ (-0.10)}} & 
\multicolumn{1}{|p{1.52cm}}{\centering
{\scriptsize code red \\ (-0.06)}} & 
\multicolumn{1}{|p{1.52cm}}{\centering
{\scriptsize cough \\ (-0.25)}} & 
\multicolumn{1}{|p{1.52cm}}{\centering
{\scriptsize cough \\ (0.13)}} & 
\multicolumn{1}{|p{1.52cm}}{\centering
{\scriptsize shortness of breath \\ (0.04)}} & 
\multicolumn{1}{|p{1.52cm}|}{\centering
{\scriptsize asthma \\ (0.06)}} \\ 
\hline
\multicolumn{1}{|p{1.52cm}}{\centering
{\scriptsize traffic \\ (0.14)}} & 
\multicolumn{1}{|p{1.52cm}}{\centering
{\scriptsize headache \\ (-0.14)}} & 
\multicolumn{1}{|p{1.52cm}}{\centering
{\scriptsize organic carbon \\ (-0.10)}} & 
\multicolumn{1}{|p{1.52cm}}{\centering
{\scriptsize wildfires \\ (0.15)}} & 
\multicolumn{1}{|p{1.52cm}}{\centering
{\scriptsize respiratory infection \\ (-0.09)}} & 
\multicolumn{1}{|p{1.52cm}}{\centering
{\scriptsize coughing \\ (0.06)}} & 
\multicolumn{1}{|p{1.52cm}}{\centering
{\scriptsize wildfires \\ (0.19)}} & 
\multicolumn{1}{|p{1.52cm}}{\centering
{\scriptsize power plants \\ (0.09)}} & 
\multicolumn{1}{|p{1.52cm}}{\centering
{\scriptsize heart murmur \\ (0.04)}} & 
\multicolumn{1}{|p{1.52cm}|}{\centering
{\scriptsize organic carbon \\ (0.05)}} \\ 
\hline
\multicolumn{1}{|p{1.52cm}}{\centering
{\scriptsize respiratory infection \\ (-0.12)}} & 
\multicolumn{1}{|p{1.52cm}}{\centering
{\scriptsize wildfires \\ (0.13)}} & 
\multicolumn{1}{|p{1.52cm}}{\centering
{\scriptsize respiratory infection \\ (-0.09)}} & 
\multicolumn{1}{|p{1.52cm}}{\centering
{\scriptsize coughing \\ (-0.14)}} & 
\multicolumn{1}{|p{1.52cm}}{\centering
{\scriptsize wheezing \\ (-0.07)}} & 
\multicolumn{1}{|p{1.52cm}}{\centering
{\scriptsize smog \\ (0.05)}} & 
\multicolumn{1}{|p{1.52cm}}{\centering
{\scriptsize wheezing \\ (-0.15)}} & 
\multicolumn{1}{|p{1.52cm}}{\centering
{\scriptsize nitrogen dioxide \\ (0.08)}} & 
\multicolumn{1}{|p{1.52cm}}{\centering
{\scriptsize tailpipe \\ (0.04)}} & 
\multicolumn{1}{|p{1.52cm}|}{\centering
{\scriptsize wildfires \\ (0.05)}} \\ 
\hline
\end{tabular}
\end{adjustbox}
\caption{Top five correlated search terms for O\textsubscript{3 }pollution in 10 U.S. cities: Jan. 1, 2010 to Dec 31, 2019.}
\end{table}

search volume data indeed provides a useable alternative to pollution monitors, with only 1.6$\%$ and 0.14$\%$ degradation in accuracy, respectively. Besides, the differences in model performance across different cities indicate that the online search pattern could vary from city to city. As shown in (Table 7), the top five correlated terms differ across US cities in 10 years. The variation of search patterns could lead to a degraded prediction performance for certain areas, leaving promising directions for improvements. \\ 

\subsubsection*{Sensitivity Analysis of Air Pollution Thresholds}
Classification thresholds play an essential role in our model. In this study, a standard deviation threshold from the mean of corresponding pollutants was used as a "probability threshold" to detect air pollution on a spatial-temporal resolution. However, the proposed method is sensitive to the threshold. We further investigate the performance of the proposed method on a variety of fixed classification thresholds. As shown in (Figure 5), (Figure 6) and (Figure 7), we fixed the classification thresholds for all ten cities for detecting Ozone, NO\textsubscript{2} and PM\textsubscript{2.5} pollutions. The result shows that the meteorological and search data are complementary and combining search and meteorological data leads to better prediction performance for all classification thresholds.

\begin{figure}[h]
\includegraphics[width=7.44cm,height=7.44cm]{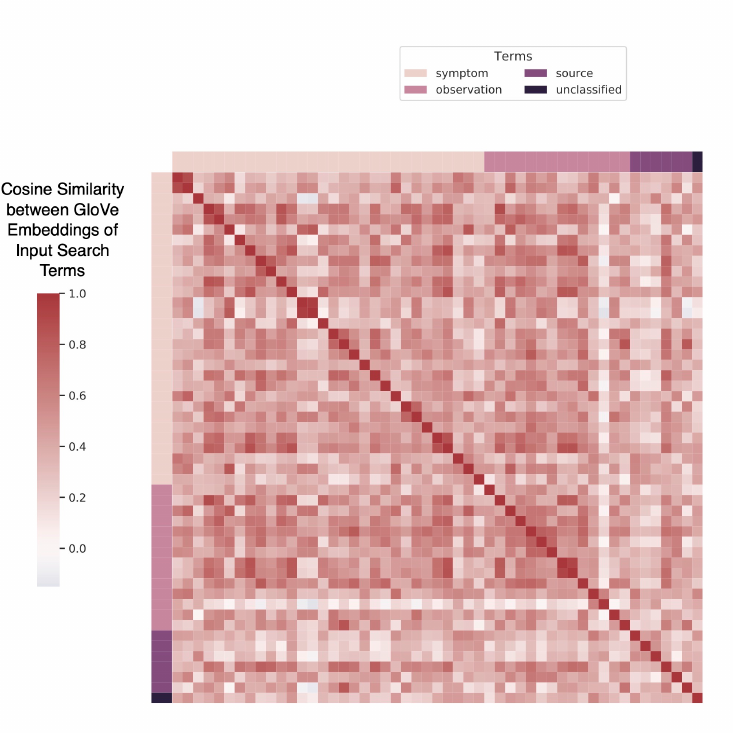}
\caption{Cosine similarity between GloVe embeddings of seed search terms.}
\end{figure}
\begin{figure}[h]
\includegraphics[width=7.44cm,height=7.44cm]{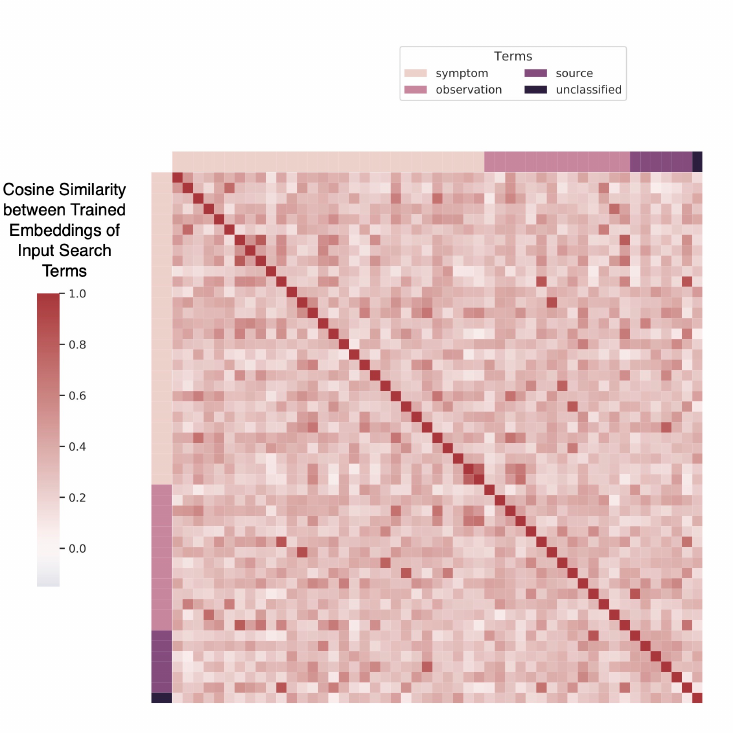}
\caption{Cosine similarity between trained embeddings of seed search terms.}
\end{figure}

\subsection*{Discussion}

\subsubsection*{Principal Findings}

In this study, we explored various existing air pollution prediction models and found that the use of a time-series neural network approach achieves the highest predictive accuracy in most of our experiments. The results showed that the LSTM-based models achieve superior accuracy for three air pollutants when both meteorological data and Web search data exist. Furthermore, our results on the inclusion of Web search data with meteorological data indicate that under short reporting delays, the LSTM models could provide highly accurate predictions compared to baseline models using meteorological and/or historical pollution concentration data. 

\begin{figure}[h]
\includegraphics[width=12.0cm,height=11.66cm]{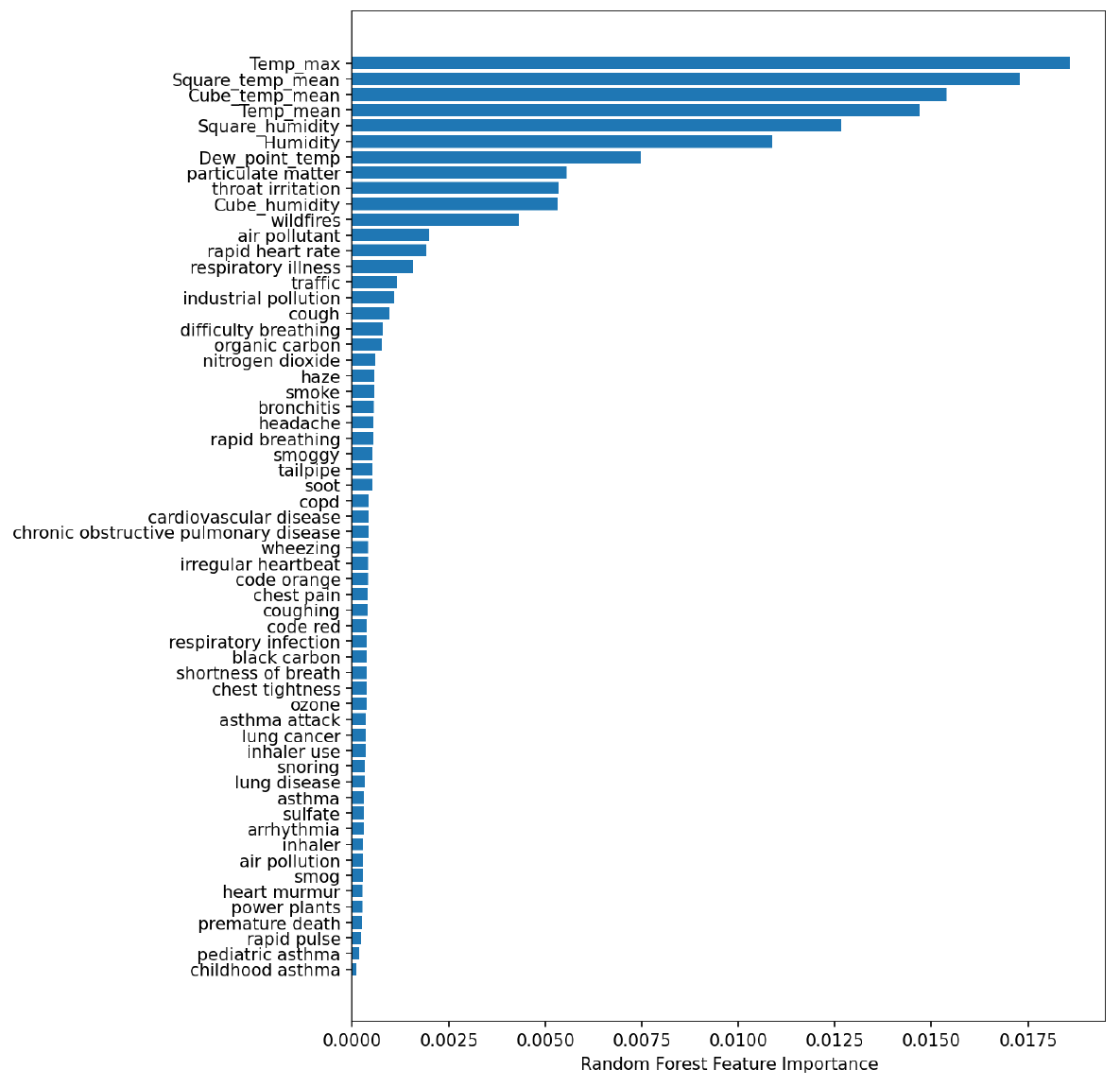}
\caption{Average feature importance for detecting Ozone pollution using Met + Search  (RF) model.}
\end{figure}

Compared to existing studies which predict urban air pollution concentrations using linear and non-linear machine learning models [19-25], our proposed method could predict air pollution when source emissions and remotely sensed satellite data are infeasible (e.g., sensed satellite data often suffers from a high missing rate due to frequent cloud cover [27]). Previous studies using online search behavior has emphasized the usage of Google Trends [30, 31] and apply regularized linear regression on colinear Web search queries to estimate disease rates from social media or online search data [28, 29, 32, 33]. Our research further explores the possibility of using LSTM models with semantic embeddings of search queries for predicting air pollution. As shown in (Figure 8) and (Figure 9), the semantic embeddings of search terms fine-tuned by the DL-LSTM model are less correlated compared to their initial GloVe embeddings, which shows that the co-linearity between search terms is reduced during the training process. 

\vspace{1\baselineskip}
We also explored the various combination of search terms and found that a comprehensive set of user queries is critical for capturing people’s responses to urban air pollution accurately. In this study, we expanded the initial set of seed terms through semantic and temporal correlations with search queries from Google Correlate. We investigate the contribution of different search term groups by manually classifying the search terms into four categories, where unclassified category includes terms with ambiguous meaning. (Table 8) shows the accuracy and F1 score when we remove search terms by categories for predicting ozone, NO\textsubscript{2},\textsubscript{ } and PM\textsubscript{2.5} pollution. Removing the search terms in categories of symptom, observation, and source categories leads to a decrease in the accuracy score for detecting at least two pollutants. At the same time, removing the search terms with ambiguous meaning only leads to a slightly higher accuracy score for all three pollutants.

\begin{figure}[h]
\includegraphics[width=12.0cm,height=11.66cm]{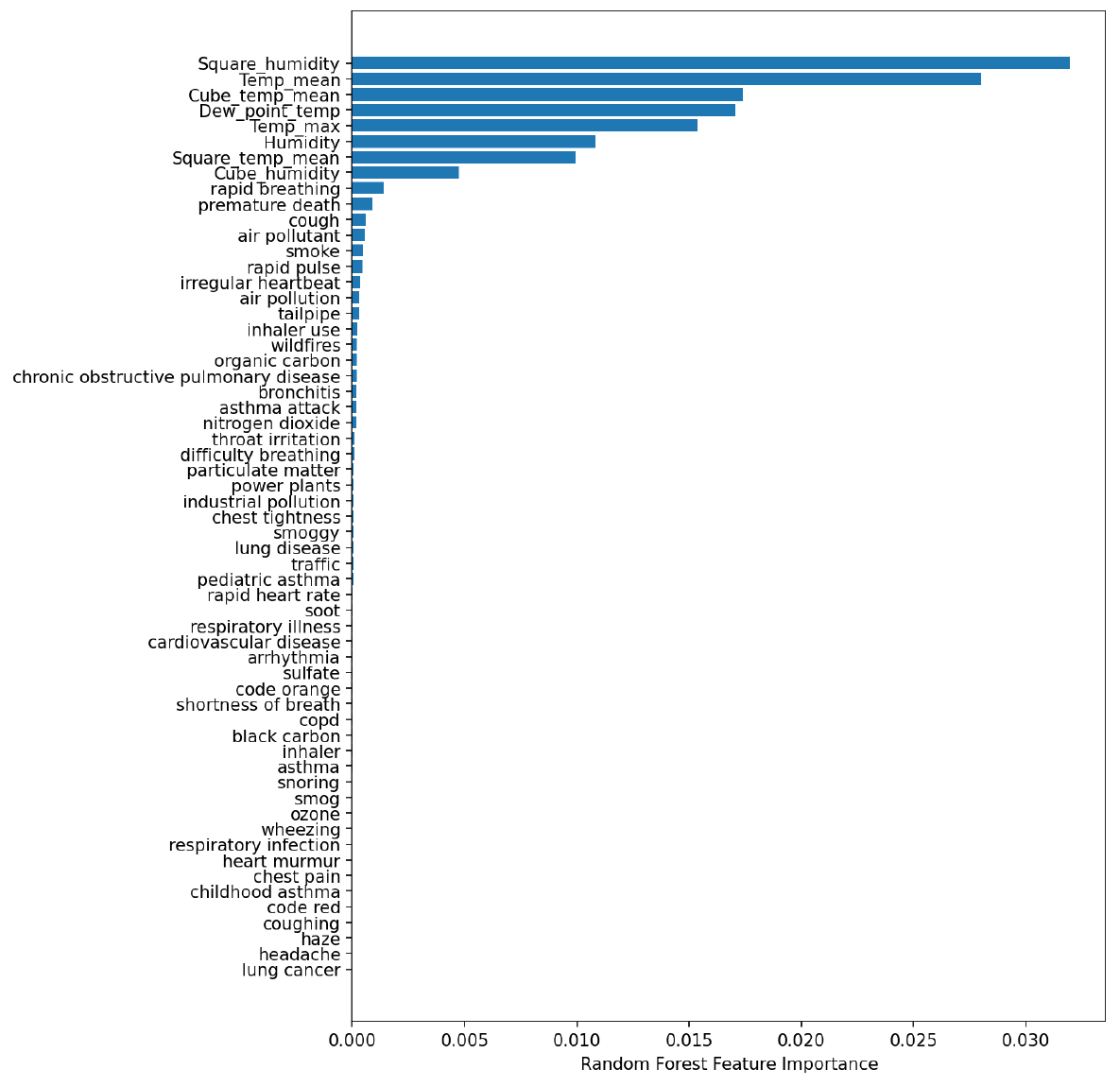}
\caption{Average feature importance for detecting NO\textsubscript{2} pollution using Met + Search  (RF) model.}
\end{figure}

By analyzing the coefficients of each search term, the result shows that several search terms contribute more than other search terms. We calculate the average feature importance of seed search terms using the RF model. As shown in (Figure 10), (Figure 11) and (Figure 12), search terms including ``particular matter", ``rapid breathing" and ``throat irritation" gain relatively high feature importance for detecting ozone, NO\textsubscript{2},\textsubscript{ }and PM\textsubscript{2.5 }pollution respectively. The result also indicates that no search terms work best for all three pollutants. 

\subsubsection*{Limitations}

A key limitation to this study is the tuning of the neural network model. First the performance of neural network models is sensitive to several hyperparameters, including optimization choices, depth, width, and regularization. Due to computational limits, we adopt a simple LSTM architecture with a single, 128-unit hidden layer and tune the model using validation datasets for other hyperparameters. In addition, we notice that the stochastic components such as the random seed for the RF model and the randomness in the optimization process of LSTM models would influence the interpretation of the results. Therefore, we repeat the experiments ten times with different random seeds for RF and LSTM models. Because the time cost of repeating LSTM models is high, we only repeat the RF, LSTM and DL-LSTM models ten times for predicting ozone pollution with all input features. The accuracy for DL-LSTM model is 0.8744 (SD 0.0046). Compared to the LSTM model (0.8714 (SD 0.0036)), the improvement is not significant (P$=$0.11). Compared to the RF model (0.8273 (SD 0.0017)), the improvement is significant (P<.001). The F1 score for DL-LSTM model is 0.6314 (SD 0.0058). Compared to both LSTM (0.6019 (SD 0.0096)) and RF model (0.5588 (SD 0.0024)), the improvements are significant (P<.001), which shows that the results of LSTM models are stable.  There is room for further exploration of more sophisticated neural network model architectures for non-infectious disease prediction. We leave the exploration of deeper and wider architecture as future work.

\begin{figure}[h]
\includegraphics[width=12.0cm,height=11.66cm]{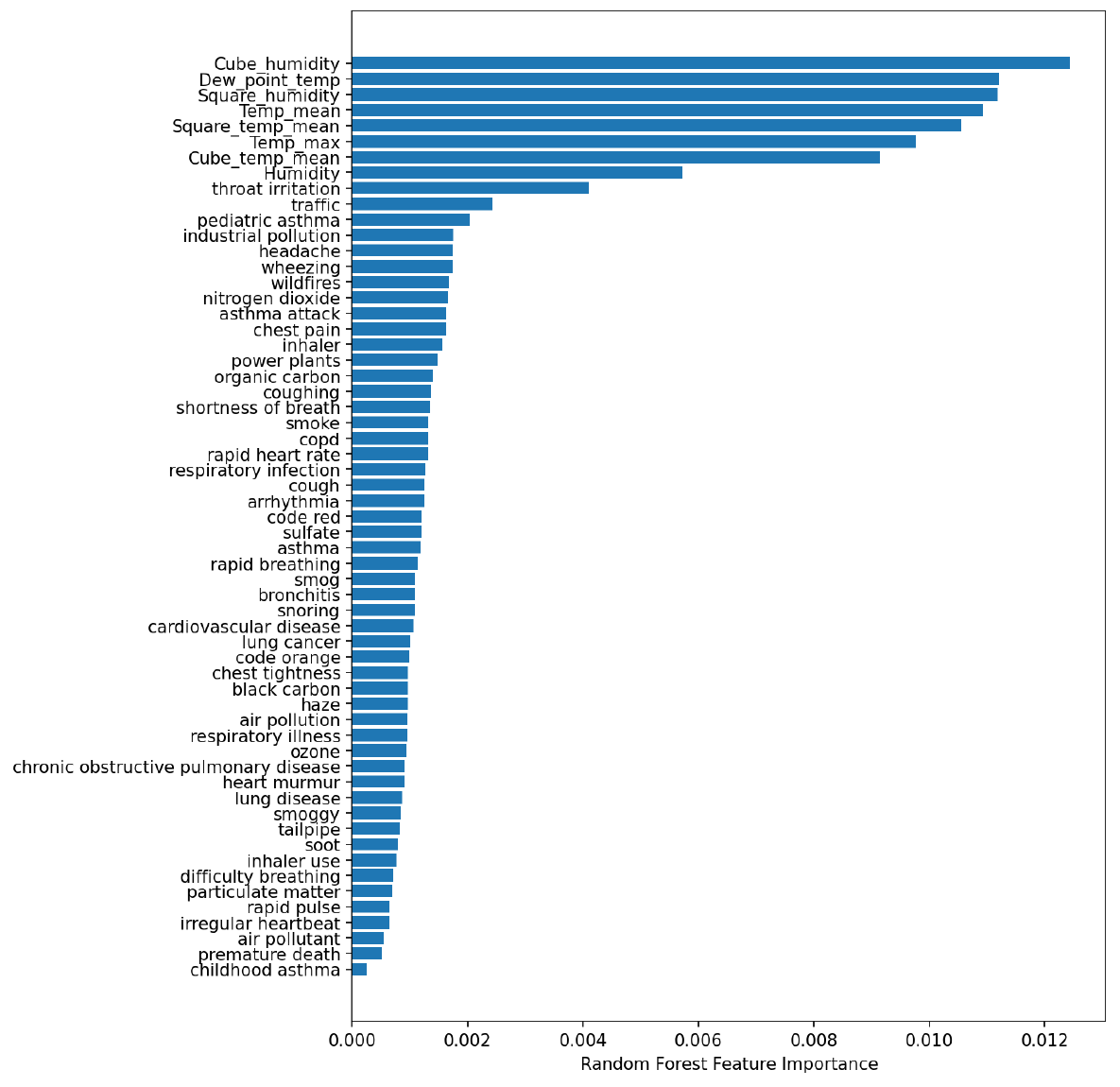}
\caption{Average feature importance for detecting NO\textsubscript{2} pollution using Met + Search  (RF) model.}
\end{figure}

Another limitation relates to the biases introduced by relying on search data, which may not reflect the underlying population demographics or experiences. While some of these issues are alleviated automatically by training a model against ground sensor pollution levels, understanding and correcting for these data biases requires further study. In the future, we plan to investigate other sources of crowd-based surveillance data such as self-reports in social media, to augment traditional physical sensor methods, thus providing a more direct, human-centered measure of how people experience elevated air pollution levels.  

\subsubsection*{Conclusions}

In this study, we posit that while Web search data cannot yet replace ground-based pollution monitors completely, it may already serve as a valuable additional signal to augment the ground-based pollution data, providing significant accuracy improvements for detecting unusual spikes in air pollution. We also find that the correlation between search terms and pollution concentration varies at the city level. Therefore, the model needs to be fine-tuned when applied to specific cities. For model and search term selection, we use the simplest LSTM architecture with a dictionary learner module, and we find that no search terms work best for all three pollutants. We propose to use our model to learn semantic correlations between available search terms to obtain better prediction results. 


\begin{table}[H]
\begin{adjustbox}{max width=\textwidth}
\begin{tabular}{p{3.8cm}p{3.8cm}p{3.8cm}p{3.8cm}p{3.8cm}p{3.8cm}p{3.8cm}p{3.8cm}}
\hline
\multicolumn{1}{|p{3.8cm}}{Pollutant } & 
\multicolumn{1}{|p{3.8cm}}{Terms} & 
\multicolumn{1}{|p{3.8cm}}{Accuracy} & 
\multicolumn{1}{|p{3.8cm}|}{F1} \\ 
\hline
\multicolumn{1}{|p{3.8cm}}{\multirow{5}{*}{\parbox{3.8cm}{\centering
Ozone}}} & 
\multicolumn{1}{|p{3.8cm}}{All} & 
\multicolumn{1}{|p{3.8cm}}{0.6961} & 
\multicolumn{1}{|p{3.8cm}|}{0.3244} \\ 
\hhline{~---}
\multicolumn{1}{|p{3.8cm}}{} & 
\multicolumn{1}{|p{3.8cm}}{All wo symptom} & 
\multicolumn{1}{|p{3.8cm}}{0.647 (-7.1$\%$)} & 
\multicolumn{1}{|p{3.8cm}|}{0.3024 (-6.8$\%$)} \\ 
\hhline{~---}
\multicolumn{1}{|p{3.8cm}}{} & 
\multicolumn{1}{|p{3.8cm}}{All wo observation} & 
\multicolumn{1}{|p{3.8cm}}{0.622 (-10.6$\%$)} & 
\multicolumn{1}{|p{3.8cm}|}{0.3264 (+0.6$\%$)} \\ 
\hhline{~---}
\multicolumn{1}{|p{3.8cm}}{} & 
\multicolumn{1}{|p{3.8cm}}{All wo source} & 
\multicolumn{1}{|p{3.8cm}}{0.6712 (-3.6$\%$)} & 
\multicolumn{1}{|p{3.8cm}|}{0.3033 (-6.5$\%$)} \\ 
\hhline{~---}
\multicolumn{1}{|p{3.8cm}}{} & 
\multicolumn{1}{|p{3.8cm}}{All wo unclassified} & 
\multicolumn{1}{|p{3.8cm}}{0.7057 (+1.4$\%$)} & 
\multicolumn{1}{|p{3.8cm}|}{0.3273 (+0.9$\%$)} \\ 
\hline
\multicolumn{1}{|p{3.8cm}}{\multirow{5}{*}{\parbox{3.8cm}{\centering
NO\textsubscript{2}}}} & 
\multicolumn{1}{|p{3.8cm}}{All} & 
\multicolumn{1}{|p{3.8cm}}{0.5684 } & 
\multicolumn{1}{|p{3.8cm}|}{0.2770} \\ 
\hhline{~---}
\multicolumn{1}{|p{3.8cm}}{} & 
\multicolumn{1}{|p{3.8cm}}{All wo symptom} & 
\multicolumn{1}{|p{3.8cm}}{0.4452 (-22.0$\%$)} & 
\multicolumn{1}{|p{3.8cm}|}{0.2418 (-12.7$\%$)} \\ 
\hhline{~---}
\multicolumn{1}{|p{3.8cm}}{} & 
\multicolumn{1}{|p{3.8cm}}{All wo observation} & 
\multicolumn{1}{|p{3.8cm}}{0.6125 (+7.8$\%$)} & 
\multicolumn{1}{|p{3.8cm}|}{0.2480 (-10.5$\%$)} \\ 
\hhline{~---}
\multicolumn{1}{|p{3.8cm}}{} & 
\multicolumn{1}{|p{3.8cm}}{All wo source} & 
\multicolumn{1}{|p{3.8cm}}{0.5452 (-4.1$\%$)} & 
\multicolumn{1}{|p{3.8cm}|}{0.2647 (-4.4$\%$)} \\ 
\hhline{~---}
\multicolumn{1}{|p{3.8cm}}{} & 
\multicolumn{1}{|p{3.8cm}}{All wo unclassified} & 
\multicolumn{1}{|p{3.8cm}}{0.6534 (+15.0$\%$)} & 
\multicolumn{1}{|p{3.8cm}|}{0.2134 (-23.0$\%$)} \\ 
\hline
\multicolumn{1}{|p{3.8cm}}{\multirow{5}{*}{\parbox{3.8cm}{\centering
PM\textsubscript{2.5}}}} & 
\multicolumn{1}{|p{3.8cm}}{All} & 
\multicolumn{1}{|p{3.8cm}}{0.8759 } & 
\multicolumn{1}{|p{3.8cm}|}{0.0699} \\ 
\hhline{~---}
\multicolumn{1}{|p{3.8cm}}{} & 
\multicolumn{1}{|p{3.8cm}}{All wo symptom} & 
\multicolumn{1}{|p{3.8cm}}{0.7897 (-9.8$\%$)} & 
\multicolumn{1}{|p{3.8cm}|}{0.1029 (+47.2$\%$)} \\ 
\hhline{~---}
\multicolumn{1}{|p{3.8cm}}{} & 
\multicolumn{1}{|p{3.8cm}}{All wo observation} & 
\multicolumn{1}{|p{3.8cm}}{0.7496 (-14.4$\%$)} & 
\multicolumn{1}{|p{3.8cm}|}{0.1049 (+50.1$\%$)} \\ 
\hhline{~---}
\multicolumn{1}{|p{3.8cm}}{} & 
\multicolumn{1}{|p{3.8cm}}{All wo source} & 
\multicolumn{1}{|p{3.8cm}}{0.8994 (+2.7$\%$)} & 
\multicolumn{1}{|p{3.8cm}|}{0.0393 (-43.8$\%$)} \\ 
\hhline{~---}
\multicolumn{1}{|p{3.8cm}}{} & 
\multicolumn{1}{|p{3.8cm}}{All wo unclassified} & 
\multicolumn{1}{|p{3.8cm}}{0.8991 (+2.6$\%$)} & 
\multicolumn{1}{|p{3.8cm}|}{0.0264 (-62.2$\%$)} \\ 
\hline
\end{tabular}
\end{adjustbox}
\caption{Accuracy and F1 score of removing different categories of search terms for detecting Ozone, NO\textsubscript{2}, PM\textsubscript{2.5} pollution using search (DL-LSTM w/ STE) as features.}
\end{table}

\vspace{12\baselineskip}


\subsubsection*{Acknowledgements}

This work was supported by grants from the US National Institutes of Health (NIH) National Library of Medicine (R21LM013014). The funders had no role in study design, data collection and analysis, decision to publish, or preparation of the manuscript. 

\subsubsection*{Conflicts of Interest}

None declared.

\subsubsection*{Abbreviations}

MSAs: Metropolitan Statistical Areas 

LSTM: Long-short Term Memory 

DL-LSTM: Dictionary Learner-Long-Short Term Memory 
\newpage

\subsection*{References}

\begin{flushleft}
{\large 1.\ \ \ \ Brynjolfsson, E., T. Geva, and S. Reichman, Crowd-Squared: Amplifying the Predictive Power of Search Trend Data. MIS Quarterly (Forthcoming), 2015. [doi: 10.25300/MISQ/2016/40.4.07]\par}
\end{flushleft}

\begin{flushleft}
{\large 2.\ \ \ \ Fung, I.C.-H., Z.T.H. Tse, and K.-W. Fu, The use of social media in public health surveillance. Western Pacific surveillance and response journal: WPSAR, 2015. 6(2): p. 3. [doi: 10.5365/wpsar.2015.6.1.019]\par}
\end{flushleft}

\begin{flushleft}
{\large 3.\ \ \ \ Hill, S., R. Merchant, and L. Ungar, Lessons learned about public health from online crowd surveillance. Big Data, 2013. 1(3): p. 160-167.\par}
\end{flushleft}

\begin{flushleft}
{\large 4.\ \ \ \ Broniatowski, D.A., M.J. Paul, and M. Dredze, National and local influenza surveillance through Twitter: an analysis of the 2012-2013 influenza epidemic. PloS one, 2013. 8(12): p. e83672.\par}
\end{flushleft}

\begin{flushleft}
{\large 5.\ \ \ \ Santillana, M., et al., Combining search, social media, and traditional data sources to improve influenza surveillance. PLoS computational biology, 2015. 11(10): p. e1004513.\par}
\end{flushleft}

\begin{flushleft}
{\large 6.\ \ \ \ Kandula, S., D. Hsu, and J. Shaman, Subregional nowcasts of seasonal influenza using search trends. Journal of medical Internet research, 2017. 19(11): p. e370.\par}
\end{flushleft}

\begin{flushleft}
{\large 7.\ \ \ \ Ning, S., S. Yang, and S. Kou, Accurate regional influenza epidemics tracking using Internet search data. Scientific reports, 2019. 9(1): p. 1-8.\par}
\end{flushleft}

\begin{flushleft}
{\large 8.\ \ \ \ Fung, I.C.-H., et al., Ebola and the social media. 2014.}
\end{flushleft}

\begin{flushleft}
{\large 9.\ \ \ \ Chan, E.H., et al., Using web search query data to monitor dengue epidemics: a new model for neglected tropical disease surveillance. PLoS neglected tropical diseases, 2011. 5(5): p. e1206.\par}
\end{flushleft}

\begin{flushleft}
{\large 10.\ \ \ \ Ayyoubzadeh, S.M., et al., Predicting COVID-19 incidence through analysis of google trends data in iran: data mining and deep learning pilot study. JMIR Public Health and Surveillance, 2020. 6(2): p. e18828. [doi: 10.2196/18828]\par}
\end{flushleft}

\begin{flushleft}
{\large 11.\ \ \ \ Fung, I.C.-H., Z.T.H. Tse, and K.-W. Fu, The use of social media in public health surveillance. Western Pacific surveillance and response journal, 2015. [doi: 10.5365/wpsar.2015.6.1.019]\par}
\end{flushleft}

\begin{flushleft}
{\large 12.\ \ \ \ De Nazelle, A., et al., Improving estimates of air pollution exposure through ubiquitous sensing technologies. Environmental Pollution, 2013. [\href{https://app.readcube.com/}{doi: 10.1016/j.envpol.2012.12.032}\href{https://app.readcube.com/}{]}\href{https://app.readcube.com/}{ }\par}
\end{flushleft}

\begin{flushleft}
{\large 13.\ \ \ \ Devarakonda, S., et al. Real-time air quality monitoring through mobile sensing in metropolitan areas. in Proc. of SIGKDD international workshop on urban computing. 2013. [\href{https://app.readcube.com/}{doi: 10.1145/2505821.2505834}\href{https://app.readcube.com/}{]}\href{https://app.readcube.com/}{ }\par}
\end{flushleft}

\begin{flushleft}
{\large 14.\ \ \ \ Snik, F., et al., Mapping atmospheric aerosols with a citizen science network of smartphone spectropolarimeters. Geophysical Research Letters, 2014. 41(20): p. 7351-7358. [\href{https://app.readcube.com/}{doi: 10.1002/2014GL061462}\href{https://app.readcube.com/}{]}\href{https://app.readcube.com/}{ }\par}
\end{flushleft}

\begin{flushleft}
{\large 15.\ \ \ \ Cohen, A.J., et al., Estimates and 25-year trends of the global burden of disease attributable to ambient air pollution: an analysis of data from the Global Burden of Diseases Study 2015. The Lancet, 2017. 389(10082): p. 1907-1918. [\href{https://app.readcube.com/}{doi: 10.1016/s0140-6736(17)30505-6}\href{https://app.readcube.com/}{]}\href{https://app.readcube.com/}{ }\par}
\end{flushleft}

\begin{flushleft}
{\large 16.\ \ \ \ Zeger, S.L., et al., Exposure measurement error in time-series studies of air pollution: concepts and consequences. Environmental health perspectives, 2000. 108(5): p. 419-426. \par}
\end{flushleft}

\begin{flushleft}
{\large 17.\ \ \ \ Sarnat, S.E., et al., Application of alternative spatiotemporal metrics of ambient air pollution exposure in a time-series epidemiological study in Atlanta. Journal of Exposure Science and Environmental Epidemiology, 2013. 23(6): p. 593. [doi: 10.1038/jes.2013.41]\par}
\end{flushleft}

\begin{flushleft}
{\large 18.\ \ \ \ Liang, D., et al., Errors associated with the use of roadside monitoring in the estimation of acute traffic pollutant-related health effects. Environmental Research, 2018. 165: p. 210-219. [\href{https://app.readcube.com/}{doi: 10.1016/j.envres.2018.04.013] }\par}
\end{flushleft}

\begin{flushleft}
{\large 19.\ \ \ \ Rybarczyk, Y. and R. Zalakeviciute, Machine learning approaches for outdoor air quality modelling: A systematic review. Applied Sciences, 2018. 8(12): p. 2570.\par}
\end{flushleft}

\begin{flushleft}
{\large 20.\ \ \ \ Chen, S., et al., Investigating China's Urban Air Quality Using Big Data, Information Theory, and Machine Learning. Polish Journal of Environmental Studies, 2018. 27(2). [doi: 10.15244/pjoes/75159]\par}
\end{flushleft}

\begin{flushleft}
{\large 21.\ \ \ \ Zhao, X., et al., A deep recurrent neural network for air quality classification. J. Inf. Hiding Multimed. Sig. Proc., 2018. 9: p. 346-354.\par}
\end{flushleft}

\begin{flushleft}
{\large 22.\ \ \ \ Lin, Y., et al. Exploiting spatiotemporal patterns for accurate air quality forecasting using deep learning. in Proc. of ACM SIGSPATIAL. 2018. [doi: 10.1145/3274895.3274907] \par}
\end{flushleft}

\begin{flushleft}
{\large 23.\ \ \ \ Zhu, W., et al., Short-term effects of air pollution on lower respiratory diseases and forecasting by the group method of data handling. Atmospheric environment, 2012. 51: p. 29-38. [doi: 10.1016/j.atmosenv.2012.01.051] \par}
\end{flushleft}

\begin{flushleft}
{\large 24.\ \ \ \ Zhang, Y., et al., Real-time air quality forecasting, part I: History, techniques, and current status. Atmospheric Environment, 2012. 60: p. 632-655.\par}
\end{flushleft}

\begin{flushleft}
{\large 25.\ \ \ \ Brokamp, C., et al., Exposure assessment models for elemental components of particulate matter in an urban environment: A comparison of regression and random forest approaches. Atmospheric Environment, 2017. 151: p. 1-11.\par}
\end{flushleft}

\begin{flushleft}
{\large 26.\ \ \ \ Cabaneros, S.M., J.K. Calautit, and B.R. Hughes, A review of artificial neural network models for ambient air pollution prediction. Environmental Modelling $\&$ Software, 2019. 119: p. 285-304. [doi: 10.1016/j.envsoft.2019.06.014]\par}
\end{flushleft}

\begin{flushleft}
{\large 27.\ \ \ \ Misra, P. and W. Takeuchi, Assessing Population Sensitivity to Urban Air Pollution Using Google Trends and Remote Sensing Datasets. The International Archives of Photogrammetry, Remote Sensing and Spatial Information Sciences, 2020. 42: p. 93-100. [doi: 10.5194/isprs-archives-xlii-3-w11-93-2020]\par}
\end{flushleft}

\begin{flushleft}
{\large 28.\ \ \ \ Zou, B., V. Lampos, and I. Cox. Transfer Learning for Unsupervised Influenza-like Illness Models from Online Search Data. in Proc. of WWW. 2019. [doi: 10.1145/3308558.3313477] \par}
\end{flushleft}

\begin{flushleft}
{\large 29.\ \ \ \ Yang, S., M. Santillana, and S.C. Kou, Accurate estimation of influenza epidemics using Google search data via ARGO. PNAS, 2015. 112(47): p. 14473-14478. [doi: 10.1073/pnas.1515373112] \par}
\end{flushleft}

\begin{flushleft}
{\large 30.\ \ \ \ Jun, S.-P., H.S. Yoo, and S. Choi, Ten years of research change using Google Trends: From the perspective of big data utilizations and applications. Technological Forecasting and Social Change, 2018. 130: p. 69-87. [doi: 10.1016/j.techfore.2017.11.009] \par}
\end{flushleft}

\begin{flushleft}
{\large 31.\ \ \ \ Carri$\textbackslash$`e, r.-S., Yan and Labb$\textbackslash$'e, Felipe, Nowcasting with Google Trends in an emerging market. Journal of Forecasting, 2013. 32(4): p. 289-298. [doi: 10.1002/for.1252] \par}
\end{flushleft}

\begin{flushleft}
{\large 32.\ \ \ \ Lampos, V. and N. Cristianini. Tracking the flu pandemic by monitoring the social web. in 2010 2nd international workshop on cognitive information processing. 2010. [doi: 10.1109/CIP.2010.5604088] \par}
\end{flushleft}

\begin{flushleft}
{\large 33.\ \ \ \ Lampos, V., et al., Advances in nowcasting influenza-like illness rates using search query logs. Scientific reports, 2015. [doi: 10.1038/srep12760] \par}
\end{flushleft}

\begin{flushleft}
{\large 34.\ \ \ \ Lampos, V., T. De Bie, and N. Cristianini. Flu detector-tracking epidemics on Twitter. in Proc. of ECML. 2010. [doi: 10.1007/978-3-642-15939-8\_42] \par}
\end{flushleft}

\begin{flushleft}
{\large 35.\ \ \ \ Lampos, V., Preo$\textbackslash$ct, iuc-Pietro, Daniel, and T. Cohn. A user-centric model of voting intention from Social Media. in Proceedings of $\{$ACL$\}$. 2013.\par}
\end{flushleft}

\begin{flushleft}
{\large 36.\ \ \ \ Zou, H. and T. Hastie, Regularization and variable selection via the elastic net. Journal of the royal statistical society: series B (statistical methodology), 2005. 67(2): p. 301-320. [doi: 10.1111/j.1467-9868.2005.00503.x] \par}
\end{flushleft}

\begin{flushleft}
{\large 37.\ \ \ \ Hochreiter, S. and J.u. Schmidhuber, rgen, Long short-term memory. Neural computation, 1997. 9(8): p. 1735-1780. [doi: 10.1162/neco.1997.9.8.1735] \par}
\end{flushleft}

\begin{flushleft}
{\large 38.\ \ \ \ Elman, J.L., Distributed representations, simple recurrent networks, and grammatical structure. Machine learning, 1991. 7(2-3): p. 195-225. [doi: 10.1007/BF00114844] \par}
\end{flushleft}

\begin{flushleft}
{\large 39.\ \ \ \ Wu, Y., et al., Google's neural machine translation system: Bridging the gap between human and machine translation. arXiv preprint arXiv:1609.08144, 2016. [arxiv: 1609.08144]\par}
\end{flushleft}

\begin{flushleft}
{\large 40.\ \ \ \ Graves, A. and N. Jaitly. Towards end-to-end speech recognition with recurrent neural networks. in ICML. 2014.\par}
\end{flushleft}

\begin{flushleft}
{\large 41.\ \ \ \ Di, Q., et al., Air pollution and mortality in the Medicare population. New England Journal of Medicine, 2017. 376(26): p. 2513-2522. [doi: 10.1056/NEJMc1709849] \par}
\end{flushleft}

\begin{flushleft}
{\large 42.\ \ \ \ Vedal, S., et al., Air pollution and daily mortality in a city with low levels of pollution. Environmental health perspectives, 2003. 111(1): p. 45-52. [doi: 10.1289/ehp.5276] \par}
\end{flushleft}

\begin{flushleft}
{\large 43.\ \ \ \ Sarnat, J.A., et al., Spatiotemporally resolved air exchange rate as a modifier of acute air pollution-related morbidity in Atlanta. Journal of Exposure Science and Environmental Epidemiology, 2013. 23(6): p. 606. [doi: 10.1038/jes.2013.32] \par}
\end{flushleft}

44.\ \ \ \ Google. Google Trends URL: https://support.google.com/trends/answer/4365533?hl$=$en{\large  }

\begin{flushleft}
{\large 45.\ \ \ \ Challet, D. and A.B.H. Ayed, Do Google Trend data contain more predictability than price returns? arXiv preprint arXiv:1403.1715, 2014.\par}
\end{flushleft}

\begin{flushleft}
{\large 46.\ \ \ \ Google. Google correlate URL:https://www.google.com/trends/correlate}
\end{flushleft}

\begin{flushleft}
{\large 47.\ \ \ \ Mikolov, T., et al. Distributed Representations of Words and Phrases and Their Compositionality. in Proc. of NeurIPS. 2013.\par}
\end{flushleft}

\begin{flushleft}
{\large 48.\ \ \ \ Pilotto, L.S., et al., Respiratory effects associated with indoor nitrogen dioxide exposure in children. International journal of epidemiology, 1997. 26(4): p. 788-796. [doi: 10.1093/ije/26.4.788] \par}
\end{flushleft}

\begin{flushleft}
{\large 49.\ \ \ \ Chauhan, A.J., et al., Personal exposure to nitrogen dioxide (NO2) and the severity of virus-induced asthma in children. The Lancet, 2003. 361(9373): p. 1939-1944. [doi: 10.1016/S0140-6736(03)13582-9] \par}
\end{flushleft}

\begin{flushleft}
{\large 50.\ \ \ \ Kane, M.J., et al., Comparison of ARIMA and Random Forest time series models for prediction of avian influenza H5N1 outbreaks. BMC bioinformatics, 2014. 15(1): p. 276. [doi: 10.1186/1471-2105-15-276] \par}
\end{flushleft}

\begin{flushleft}
{\large 51.\ \ \ \ He, K., et al. Delving deep into rectifiers: Surpassing human-level performance on imagenet classification. in Proc. of CVPR. 2015. [doi: 10.1109/ICCV.2015.123]\par}
\end{flushleft}

\begin{flushleft}
{\large 52.\ \ \ \ Pennington, J., R. Socher, and C.D. Manning. Glove: Global vectors for word representation. in Proc. of the 2014 conference on empirical methods in natural language processing (EMNLP). 2014. [doi: 10.3115/v1/D14-1162] \par}
\end{flushleft}

\begin{flushleft}
{\large 53.\ \ \ \ Kreindler D M, Lumsden C J. The Effects of the Irregular Sample and Missing Data in Time Series Analysis[J]. Nonlinear dynamics, psychology, and life sciences, 2006. [doi: 10.1201/9781439820025-9]\par}
\end{flushleft}

\begin{flushleft}
{\large 54. \ \ \ \ Deng S, Wang S, Rangwala H, et al. Graph message passing with cross- \ \ \ location attentions for long-term ILI prediction[J]. arXiv preprint arXiv:1912.10202, 2019. [doi: 10.1145/3340531.3411975]\par}
\end{flushleft}

\begin{flushleft}
{\large 55. \ \ \ \ Zou B, Lampos V, Cox I. Multi-task learning improves disease models from web search[C]//proceedings of the 2018 world wide web conference. 2018: 87-96. [doi: 10.1145/3178876.3186050]\par}
\end{flushleft}

\begin{flushleft}
{\large 56.\ \ \ \ Zhang Y, Yakob L, Bonsall M B, et al. Predicting seasonal influenza epidemics using cross-hemisphere influenza surveillance data and local internet query data[J]. Scientific reports, 2019, 9(1): 1-7. [doi: 10.1038/s41598-019-39871-2]\par}
\end{flushleft}

\begin{flushleft}
{\large 57.\ \ \ \ Cohen AJ, Brauer M, Burnett R, Anderson HR, Frostad J, Estep K, Balakrishnan K, Brunekreef B, Dandona L, Dandona R, Feigin V. Estimates and 25-year trends of the global burden of disease attributable to ambient air pollution: an analysis of data from the Global Burden of Diseases Study 2015. The Lancet. 2017 May 13;389(10082):1907-18. [doi: /10.1016/S0140-6736(17)30505-6]\par}
\end{flushleft}

\begin{flushleft}
{\large 58.\ \ \ \ Kelly FJ, Fussell JC. Air pollution and public health: emerging hazards and improved understanding of risk. Environmental geochemistry and health. 2015 Aug;37(4):631-49. [doi: 10.1007/s10653-015-9720-1]\par}
\end{flushleft}

\begin{flushleft}
{\large 59.\ \ \ \ Sarnat JA, Ted RA, Liang D, Moutinho JL, Golan R, Weber RJ, Gao D, Sarnat SE, Chang HH, Greenwald R, Yu T. Developing Multipollutant Exposure Indicators of Traffic Pollution: The Dorm Room Inhalation to Vehicle Emissions (DRIVE) Study. Research Reports: Health Effects Institute. 2018;2018. \par}
\end{flushleft}

\end{document}